\documentclass[11pt]{article}

\usepackage{amsmath}
\usepackage{graphicx,psfrag,epsf}
\usepackage{enumerate}
\usepackage{biblatex}
\usepackage{amssymb}
\usepackage{comment}
\usepackage{amsthm}
\usepackage{changepage}
\usepackage{mathtools}
\usepackage{booktabs}
\usepackage{multirow}

\usepackage{algorithm}

\usepackage{commath}
\usepackage{enumitem}
\usepackage{tikz}
\usepackage{ragged2e}
\usepackage{authblk}

\usepackage{hyperref}
\usepackage{graphicx}
\usepackage{xcolor}
\usepackage{subcaption}
\usepackage{bm}
\usepackage{soul} 

\newcommand{\blind}{0}

\newcommand{\fun}{\xi}

\DeclareMathOperator*{\diag}{diag}

\DeclareMathOperator*{\E}{{\rm E}}

\newcommand{\R}{\mathbb{R}}
\newcommand{\Var}{{\rm Var}}

\newcommand{\U}{U}

\newcommand{\Rec}{{\rm Re}}

\newtheorem{theorem}{Theorem}[section]
\newtheorem{proposition}[theorem]{Proposition}

\newtheorem{definition}[theorem]{Definition}

\numberwithin{equation}{section}

\author[1]{Diogo Pinheiro}
\author[1,2,*]{M. Rosário Oliveira}
\author[2]{Igor Kravchenko}
\author[3]{Lina Oliveira}

\affil[1]{Dep. of Mathematics, Instituto Superior Técnico, ULisboa, Portugal}
\affil[2]{CEMAT, Instituto Superior Técnico, ULisboa, Portugal}
\affil[3]{CAMGSD and Dep. of Mathematics, Instituto Superior Técnico, ULisboa, Portugal}

\affil[*]{Corresponding author: 
\href{mailto:rosario.oliveira@tecnico.ulisboa.pt}{rosario.oliveira@tecnico.ulisboa.pt}}

\def\spacingset#1{\renewcommand{\baselinestretch}%
{#1}\small\normalsize} \spacingset{1}

\if0\blind
{
  \title{\bf Interval Fisher's Discriminant Analysis and Visualisation}
}

\if1\blind
{
  \bigskip
  \bigskip
  \bigskip
  \begin{center}
    {\LARGE\bf Title}
\end{center}
  \medskip
} \fi

\begin{document}

\maketitle

\begin{abstract}
In Data Science, entities are typically represented by single valued measurements. Symbolic Data Analysis extends this framework to more complex structures, such as intervals and histograms, that express internal variability. We propose an extension of multiclass Fisher's Discriminant Analysis to interval-valued data, using Moore's interval arithmetic and the Mallows' distance. Fisher's objective function is generalised to consider simultaneously the contributions of the centres and the ranges of intervals and is numerically maximised. The resulting discriminant directions are then used to classify interval-valued observations.
To support visual assessment, we adapt the class map, originally introduced for conventional data, to classifiers that assign labels through minimum distance rules. We also extend the silhouette plot to this setting and use stacked mosaic plots to complement the visual display of class assignments. Together, these graphical tools provide insight into classifier performance and the strength of class membership.
Applications to real datasets illustrate the proposed methodology and demonstrate its value in interpreting classification results for interval-valued data.
\end{abstract}

\noindent%

\spacingset{1} 

\section{Introduction}
\label{sec:intro}

In many applications, from sensor networks and finance to medicine and social science, observations are naturally recorded not as single values but as interval-valued data, capturing inner variability, possibly resulting from aggregating microscale data or may exist in their own right. Traditional classification methods, however, are largely built on the assumption of point observations and hence misrepresent or discard the intrinsic information encoded in intervals. Symbolic Data Analysis (SDA) is the field that addresses
these types of data. Intervals, histograms, and distributions are alternative data types studied in this field. 
Data expressed as multimensional intervals gives rise to a pressing methodological gap: how to perform discriminant analysis directly on interval‐valued variables in a way that respects their internal structure and yields interpretable classification rules. 

Extensions of Fisher’s Discriminant Analysis (FDA) to interval data were first proposed by \cite{Silva2006}, who introduced three formulations: one defining dispersion measures and Fisher matrices under a uniform microdata assumption whilst the other transform intervals into conventional data, either through vertex expansion or separate analyses of centres and ranges. Although valid, these transformations reduce interpretability by discarding the intrinsic structure of interval data. A subsequent parametric model by \cite{Silva2015} assumed multivariate Normal or Skew-Normal distributions for the centres and log-ranges, aligning with Linear Discriminant Analysis but departing from Fisher’s original distribution-free framework.

Later on, \cite{Dias2021} developed a two-class FDA for histogram-valued data (an extension of interval data) based on the decomposition of inertia of the squared Mallows' distance, assuming uniformly distributed microdata. 
In their formulation, the Fisher problem involves identifying $2p$ nonnegative coefficients that maximize the discriminant criterion. Our proposal generalizes this framework to multiple classes and determines discriminant directions directly in a $p$-dimensional space.

This paper proposes a new multi‐class Fisher-type discriminant framework for interval‐valued data. Our method advances over earlier work in several ways. Firstly, rather than reducing intervals to point estimates (such as centres) or expanding them into high‐dimensional conventional vectors (e.g., all hyper‐rectangle vertices), we model the intervals directly using appropriate mathematical and statistical tools to deal with intervals, and derive discriminant directions in the original interval‐space. Secondly, our formulation is built on the decomposition of inertia under the squared Mallows' distance, yielding an intuitive weighting of centres and ranges, based on general microdata distributional assumptions and not only uniform assumption. Third, unlike most existing approaches which handle only binary classification, our method accommodates multiple classes in a unified way.

We detail the theoretical development of the method, including the derivation of within‐ and between‐group matrices, the optimisation of discriminant directions, and the classification rule based on the Mallows' distance. We then demonstrate its performance via simulation studies and real‐data applications. Our results show that the proposed approach not only lead to high classification performance but also enhances interpretability by explicitly decomposing the contributions of interval‐centres and interval‐ranges in the discriminant space. We thus proposed a statistically principled, computationally feasible, and conceptually clear and interpretable solution to the multi-class classification of interval‐valued data. 

As a second major contribution, we introduce a suite of visualization tools that enhance the interpretability of classification outcomes, extending to the symbolic framework the visualization principles proposed by \cite{RaymaekersSih:2022,RaymaekersClassmap:2022} \cite{Raymaekers.Rousseuw:2024}.
The stacked mosaic plot displays, for each true class, the proportion of observations assigned to each predicted class, thereby highlighting both the percentage of misclassified cases and the classes in which these errors are most prevalent.
Building on the notion of farness, an outlyingness measure, correctly classified observations can also be identified as outliers.
The farness score is complemented by an additional measure, denoted ${\rm \ell DAC}$, which quantifies the proximity between the {true} class and its nearest alternative, thereby providing an interpretable measure of the classifier’s confidence in each assignment.
This joint visualization enables the simultaneous assessment of classification certainty and observation outlyingness across classes.
Moreover, the coefficient {${\rm \ell DAC}$} naturally leads to the definition of a silhouette coefficient, analogous to that used in cluster analysis \cite{Rousseeuw1987}, yielding new visualization tools and diagnostic indices (named class average and overall average silhouette coefficient) for evaluating classifier performance.

The article is organized as follows.
Section \ref{sec:Fundam} outlines the mathematical and statistical foundations underlying the proposed methodology.
In Section \ref{sec:IFDA}, we introduce the symbolic formulation of Fisher's problem, developed using the decomposition of inertia based on Mallows' distance, and describe the corresponding numerical optimisation procedure.
Section \ref{sec:Visualization} presents several visualization tools that further enhance the interpretability of the classification results summarized by the confusion matrix and associated performance metrics, enabling the identification of outliers.
Section \ref{sec:NumStudies} evaluates the effectiveness of the proposed approach through a simulation study and two real-world applications involving the classification of Internet traffic into regular or attacks under specific redirection relays.
Finally, Section \ref{sec:Concl} provides concluding remarks, and all technical proofs are deferred to the Supplementary Materials. 

\section{Fundamentals}
\label{sec:Fundam}

In this section, we introduce the notation, basic results, and definitions used throughout the remainder of the paper. Given a vector $\bm{v}=(v_1,\ldots,v_p)^T\in\R^p$, we define $\diag(\bm{v})$ as the diagonal matrix whose main diagonal is $\bm{v}$. Similarly, for a matrix $\bm{A}$, we define $\diag(\bm{A})$ as the diagonal matrix sharing the same main diagonal as $\bm{A}$. 
Furthermore, let sgn$(\boldsymbol{v}) = (\textnormal{sgn}(v_1),\ldots,\textnormal{sgn}(v_p))^T$ be the vector of signs of the components of $\boldsymbol{v}$, where the sign of a real number is $-1$, if it is negative, and $1$, otherwise.

\subsection{Interval-Valued Data}\label{subsec:interval}

In SDA, a symbolic interval $x$ captures intrinsic variability through the distribution $F$ of the underlying conventional points, which we refer to as microdata. These microdata are often inaccessible (latent) to the researcher due to privacy restrictions or other limitations. The aggregation of microdata yields a real-valued interval, called macrodata, which can be expressed either as $[a,b]$ ($b\geq a$) or equivalently by its centre–range representation, given by the bivariate vector $(c,r)^T$, where $c=(a+b)/2$ and $r=b-a \geq 0$. A symbolic interval is thus fully specified by its macrodata and the distribution of its microdata. Accordingly, $x$ can be succinctly represented as $x=(c,r,F)$. 

A $p$-dimensional symbolic interval $\boldsymbol{x}=(x_1,\ldots ,x_p)^T$, $p \in \mathbb{N}$, known as symbolic hyperrectangle, is also defined by its macrodata and microdata. The macrodata correspond to the $p$-dimensional real-valued interval $([a_1,b_1],\ldots,[a_p,b_p])^T$ with $b_i \geq a_i$, or, equivalently, to the centre and range vector $(\boldsymbol{c}^T,\boldsymbol{r}^T)^T$, where $\boldsymbol{c}=(c_1,\ldots,c_p)^T$, $\boldsymbol{r}=(r_1,\ldots,r_p)^T$, and $r_i \geq 0$, $i=1,\ldots,p$. For simplicity, the microdata are described by the set of marginal distributions $F_{V_1},\ldots,F_{V_p}$, which capture the internal variability within each univariate real-valued interval $[a_i,b_i]$. In concise form, we write $\boldsymbol{x}=(\boldsymbol{c},\boldsymbol{r},F_{\boldsymbol{x}})$, where 
$F_{\boldsymbol{x}}(v_1,\ldots,v_p) = \prod_{i=1}^p F_{V_i}(v_i)$, assuming independence between variables.
A conventional point $\boldsymbol{v} = (v_1,\dots,v_p)^T \in \mathbb{R}^p$ is a degenerate case where $v_i = a_i = b_i$, or, equivalently, $v_i=c_i$, $r_i=0$, and $F_{V_i}$ is the distribution function of a degenerate random variable $V_i$ satisfying $P(V_i=v_i)=1$, $i = 1,\dots,p$.

In \cite{Oliveira2021}, a model was proposed to link the macrodata and the microdata of a symbolic interval. The key idea is to introduce a latent random variable $U$ with support on $[-1,1]$ as the source of inner variability. The microdata $V$ are then obtained by scaling realisations of $U$ to the interval $[a,b]$ using the centre and the range, i.e., $V = c + U\, r/2$. The extension to hyperrectangles is analogous: each component is modelled separately by introducing a set of independent latent random variables $U_i \in [-1,1]$ with distribution $F_{U_i}$, $i=1,\ldots,p$. Under this framework, the symbolic hyperrectangle $\boldsymbol{x}=(x_1,\ldots ,x_p)^T$ is expressed as $\boldsymbol{x}=(\boldsymbol{c},\boldsymbol{r},F_{\boldsymbol{U}})$, where $\boldsymbol{U} = (U_1,\ldots,U_p)^T$ is a random vector with distribution function $F_{\boldsymbol{U}}(u_1,\ldots,u_p)= \prod_{i=1}^p F_{U_i}(u_i)$. 

\subsection{Mallows' distance}

The Mallows' distance, also known as the \mbox{$L_2$--Wasserstein} distance, is widely used in SDA. It quantifies how far apart two symbolic intervals are through the quantile functions associated with the distributions of their microdata.

\begin{definition}
Let $\boldsymbol{x}_1 = (x_{11},\dots,x_{1p})^T$ and $\boldsymbol{x}_2 = (x_{21},\dots,x_{2p})^T$  be two $p$-dimensional symbolic intervals, $p \in \mathbb{N}$, where $\boldsymbol{x}_h=(\boldsymbol{c}_h,\boldsymbol{r}_h,F_{\boldsymbol{x}_h})$, $F_{\boldsymbol{x}_h}(v_{1},\ldots,v_{p})= \prod_{i=1}^p F_{V_{hi}}(v_i)$, $h=1,2$, $i = 1,\ldots,p$. In addition, assume that $F_{V_{hi}}$ has finite second moment and let $F_{V_{hi}}^{-1}$ be the respective quantile function.
The squared Mallows' distance between $\boldsymbol{x}_1$ and $\boldsymbol{x}_2$ is defined as 
\begin{equation}\label{mallowsp}
d_M (\boldsymbol{x}_1, \boldsymbol{x}_2)^2 = \sum_{i = 1}^p{ \int_{0}^{1}{\left(F_{V_{1i}}^{-1}(t) - F_{V_{2i}}^{-1}(t)\right)^2}dt . }
\end{equation}
\end{definition}

Relying on the model proposed in \cite{Oliveira2021} and, consequently, on the formulation of a symbolic hyperrectangle $\boldsymbol{x}_h=(x_{h1},\ldots ,x_{hp})^T$ as $\boldsymbol{x}_h=(\boldsymbol{c}_h,\boldsymbol{r}_h,F_{\boldsymbol{U}_h})$, a general mathematical expression of the Mallows' distance was proved in \cite[Thm.~3.5]{Oliveira2025} by assuming that the latent random variables are absolutely continuous. 
This expression can also be extended to discrete distributions by using similar arguments and is presented in the following theorem.

\begin{theorem}\label{Th:Mallows.distance} 
Let $\boldsymbol{x}_1 = (x_{11},\dots,x_{1p})^T$ and $\boldsymbol{x}_2 = (x_{21},\dots,x_{2p})^T$  be two $p$-dimensional symbolic intervals, where $\boldsymbol{x}_h=(\boldsymbol{c}_h,\boldsymbol{r}_h,F_{\boldsymbol{U}_h})$, and
$F_{\boldsymbol{U}_h}(u_{1},\ldots,u_{p})=\prod_{i=1}^p F_{U_{hi}}(u_i)$, $h=1,2$, $i = 1,\ldots,p$. Furthermore, let ${U}_{1i}$ and ${U}_{2i}$ be identically distributed to an absolutely continuous or discrete random variable $U_i$ with support in $[-1,1]$ and finite first two moments, $\E[\U_i]$ and $\E[\U^2_i]$.
The squared Mallows' distance between $\boldsymbol{x}_1$ and $\boldsymbol{x}_2$ is given by
\begin{align}
    d_M(\boldsymbol{x}_1,\boldsymbol{x}_2)^2\ &=\ (\boldsymbol{c}_1 - \boldsymbol{c}_2)^T(\boldsymbol{c}_1 - \boldsymbol{c}_2)  + (\boldsymbol{r}_1 - \boldsymbol{r}_2)^T \boldsymbol{\Delta} (\boldsymbol{r}_1 - \boldsymbol{r}_2) \nonumber\\
    &+\ (\boldsymbol{c}_1 - \boldsymbol{c}_2)^T \boldsymbol{\Psi} (\boldsymbol{r}_1 - \boldsymbol{r}_2),
    \label{Eq:Mallows.Distance}
\end{align}
where $\bm{\Delta}=\diag\left((\delta_1,\ldots,\delta_p)^T\right)$ with $\delta_i=\E[\U_i^2]/4$, and
$\bm{\Psi}=\diag\left(\big(\E[\U_1],\ldots,\E[\U_p])^T\right)$.
\end{theorem}

Under the conditions of Theorem \ref{Th:Mallows.distance}, if the variables $U_i$, $i=1,\ldots,p$, are symmetric and identically distributed as a common random variable $U$, then $\E[U] = \E[U_i] = 0$, $\E[U^2] = \E[U_i^2] = \Var[U]$, and the Mallows' distance \eqref{Eq:Mallows.Distance} reduces to
\begin{equation}\label{Eq:mallowsdelta}
d_M(\boldsymbol{x}_1,\boldsymbol{x}_2)^2\ =\ (\boldsymbol{c}_1 - \boldsymbol{c}_2)^T(\boldsymbol{c}_1 - \boldsymbol{c}_2) + \delta (\boldsymbol{r}_1 - \boldsymbol{r}_2)^T (\boldsymbol{r}_1 - \boldsymbol{r}_2),
\end{equation}
where $\delta = \Var[U]/4 \in [0,1/4]$.

When a symbolic method is based on the Mallows' distance, the microdata are often assumed to be continuous uniformly distributed within the interval (i.e., $U \sim \mathrm{Unif}[-1,1]$). In this case, expression \eqref{Eq:mallowsdelta} applies with $\delta = 1/12$. Table~\ref{tab:1} lists a selection of well-known symmetric distributions considered in \cite{Oliveira2021} and their corresponding $\delta$ values. Here, $\mathrm{T}[-1,0,1]$ ($\mathrm{InvT}[-1,0,1]$) denotes a symmetric triangular (inverse triangular) distribution on $[-1,1]$, $\phi$ ($\Phi$) denotes the standard normal probability density (distribution) function, and $\mathcal{N}\big(0,\frac{1}{9}\big) \mid [-1,1]$ denotes a normal distribution with mean zero and variance $1/9$, truncated to $[-1,1]$. The conventional scenario is represented by a degenerate distribution with P$(U = 0) = 1$. 

\begin{table}[h!]
\small
\centering
\caption{List of coefficients $\delta$, in descending order, associated with the squared Mallows' distance \eqref{Eq:mallowsdelta} for intervals with several symmetric distributions for $U$.}
\begin{tabular}{lc}
    \toprule\addlinespace[3pt]
    $U$ Distribution & \hspace{1pt}$\hspace{-2pt}\delta = \Var[U]/4$ \\ \addlinespace[3pt]\midrule
    \addlinespace[5pt] 
    $U \sim \mathrm{Unif}\{-1,1\}$ & 1/4 \\ \addlinespace[3pt] 
    $U \sim \mathrm{InvT}[-1,0,1]$ &  1/8\\ 
    $U \sim \mathrm{Unif}[-1,1]$ & 1/12 \\ 
    $U \sim \mathrm{T}[-1,0,1]$ & 1/24\\ 
    $U \sim \mathcal{N}\big(0,\frac{1}{9}\big) | [-1,1]$& $1/36 - \phi(3)/\left({6(2\Phi(3)-1)}\right)$ \\ 
    P$(U = 0) = 1$ & $0$ \\ 
    \bottomrule
\end{tabular}
\label{tab:1}
\end{table}

When the $U_i$, $i=1,\ldots, p$, are symmetric random variables, the Mallows' distance for $p$-dimensional intervals parallels the Euclidean distance in $\R^{2p}$, with the contribution of the ranges downweighted by coefficients $\delta_i \in [0,1/4]$.  
The upper bound for $\delta_i$ corresponds to the discrete uniform distribution, where the microdata are concentrated in the endpoints of the macrodata of the symbolic interval. On the other end, the conventional case ($\delta = 0$) removes the contribution of the ranges completely, and the Mallows' distance becomes the Euclidean distance for real-valued data.

\subsection{Barycentre}
The symbolic centroid, or barycentre, is defined in \cite{Irpino2015} as the distribution that minimises the sum of squared Mallows' distances within a set of distributions. For interval-valued data, it can be shown \cite[pp.~189--190]{Irpino2006} that, under a fixed distribution of the latent random variables, the sample barycentre is the symbolic hyperrectangle whose centre and range are the averages of the centres and ranges of the constituent hyperrectangles, respectively. The proof for the population barycentre can be found in \cite[Thm.~4.1]{Oliveira2025}. This also holds for the sample barycentre, since it is a particular case by assuming that the population follows the empirical distribution over the observed centres and the ranges. The sample version is summarized in the following theorem.

\begin{theorem}
    Let $\boldsymbol{x}_1,\ldots,\boldsymbol{x}_n$ be a sample of $p$-dimensional symbolic intervals, with $p, n\in\mathbb{N}$, $\boldsymbol{x}_h = (\boldsymbol{c}_h, \boldsymbol{r}_h, F_{\boldsymbol{U}})$, $h = 1,\ldots, n$, where $F_{\boldsymbol{U}}$ is the distribution of the random vector $\boldsymbol{U} = (U_1,\ldots,U_p)^T$, consisting of latent random variables with finite second moment. The sample barycentre, $\overline{\boldsymbol{x}}_{B}$, is the $p$-dimensional symbolic interval that minimises the sum of squared Mallows' distances to all $\boldsymbol{x}_h$. That is,
\begin{align}\label{barycentre}
     \overline{\boldsymbol{x}}_{B} \coloneqq  \underset{\boldsymbol{x}\ =\ (\boldsymbol{c}, \boldsymbol{r}, F_{\boldsymbol{U}})}{\mathrm{arg}\ \mathrm{min}}\sum_{h=1}^{n}d_M(\boldsymbol{x}_h, \boldsymbol{x})^2= (\overline{\boldsymbol{c}}_n, \overline{\boldsymbol{r}}_n, F_{\boldsymbol{U}}), %
\end{align}
where $\overline{\boldsymbol{c}}_n = \sum_{h=1}^{n}\boldsymbol{c}_{h}/n$ and $\overline{\boldsymbol{r}}_n = \sum_{h=1}^{n}\boldsymbol{r}_{h}/n$ are, respectively, the average of the centres and the average of the ranges of $\boldsymbol{x}_1,\ldots,\boldsymbol{x}_n$.
\end{theorem}

In the previous result, we show that the barycentre of a set of symbolic hyperrectangles with the same distribution function $F_{\boldsymbol{U}}$ also shares this distribution. In other words, the degree of freedom relative to the microdata is removed, and we define a minimisation problem as a function of the macrodata.

\subsection{Moore's algebraic structure}

One of the most remarkable qualities of FDA is its ability to address a multidimensional classification problems by linearly projecting the original data  on a lower-dimensional space. Extending this concept to interval observations (macrodata) is essential for adapting FDA to the symbolic framework. In SDA, multiple definitions exist for the linear combination of real intervals.  
 In this work, we consider the Moore's definition \cite{Moore2009MOORE}. Other alternatives to Moore's algebraic structure can be found in \cite{Rodrigo2023}.

\begin{definition}
    Let $\boldsymbol{x} = (\boldsymbol{c}, \boldsymbol{r}, F_{\boldsymbol{U}})$ be a symbolic hyperrectangle with centre $\boldsymbol{c} = (c_1,\ldots, c_p)^T$ and range $\boldsymbol{r} = (r_1,\ldots, r_p)^T$, and $\boldsymbol{\alpha} = (\alpha_1,\ldots,\alpha_p)^T \in \R^p$, $p \in \mathbb{N}$, a real-valued vector. Then, Moore's linear combination of symbolic intervals is
\begin{equation}\label{moore}
    \boldsymbol{\alpha}^T\boldsymbol{x} = 
    \left(\sum_{i=1}^{p}{\alpha_ic_i},\sum_{i=1}^{p}{|\alpha_i|r_i}, F_{U}\right) =  \left(\boldsymbol{\alpha}^T\boldsymbol{c},|\boldsymbol{\alpha}|^T\boldsymbol{r}, F_{U} \right),
\end{equation}
where $|\boldsymbol{\alpha}| = (|\alpha_1|,\ldots,|\alpha_p|)^T$ is called the absolute value vector of $\boldsymbol{\alpha}$, and $F_U$ is a distribution function related to the latent scaled microdata in $\boldsymbol{\alpha}^T\boldsymbol{x}$.
\end{definition}

Note that Moore's structure does not provide an explicit connection between the original distribution $F_{\boldsymbol{U}}$ and the distribution function $F_U$ defined on the new projected space. Instead, the latter distribution is selected on the basis of previous knowledge or fine-tuned \textit{a posteriori} using classification results. In particular, we restrict our study to symmetric distributions such as the ones in Table~\ref{tab:1}, and the simplified Mallows' distance \eqref{Eq:mallowsdelta}.

\section{Interval Fisher's Discriminant Analysis}\label{sec:IFDA}

We adapt Fisher's methodology to the interval scenario, here denominated Interval Fisher's Discriminant Analysis (IFDA), 
where the symbolic concepts previously described their conventional counterparts. To do this, we first set the interval classification paradigm and replace required notation. 

For a fixed distribution function $F_U$, let $y_1 , \ldots, y_n$, where $y_i= (c_i,r_i, F_U)$, $i=1,\ldots, n$, $n\in\mathbb{N}$, be a sample of symbolic intervals, divided into $g$ classes, each of size $n_j$, $j=1,\ldots,g$, such that $\sum_{j=1}^g n_j = n$. In addition, let $\overline{y}_B = \sum_{h=1}^ny_h/n= (\overline{c}_y, \overline{r}_y, F_U)$ be the overall sample barycentre and $\overline{y}_j = \sum_{h \in j}y_h/n_j = (\overline{c}_{y_j}, \overline{r}_{y_j}, F_U)$ be the sample barycentre of the $j$-th class, defined according to \eqref{barycentre}. The notation $``h \in j"$ is used to represent observation $h$ in the $j$-th class. 

In \cite[190]{Irpino2006}, it was shown that the total inertia (TI) in a system of labelled symbolic intervals, measured using the sum of the squared Mallows' distances to the barycentre, could be decomposed into two parts: the inertia between the different classes in the system (BI) and the inertia within each group (WI). This decomposition of the variability of the system induced by the squared Mallows' distance is formally established by
\begin{equation}\label{inertia}
    \mathrm{TI}\ =\ \mathrm{BI}\ +\ \mathrm{WI} \ \iff\ \sum_{h=1}^n d_M(y_h, \overline{y}_B)^2\ =\ \sum_{j=1}^gn_j\hspace{1pt}d_M(\overline{y}_{j},\overline{y}_B)^2\ + \ \sum_{j=1}^g\sum_{h\in j} d_M(y_h,\overline{y}_{j})^2\ .
\end{equation}

An argument commonly used to derive conventional Fisher’s Discriminant Analysis is the decomposition of inertia of the squared Euclidean distance together with the between-scatter and within-scatter matrices associated with the classes in the original space \cite{Ghojogh2023}. We adapt this argument to the symbolic case. To simplify the expressions, we assume that $F_U$ is a symmetric distribution, which allows the use of the Mallows' distance formulation \eqref{Eq:mallowsdelta}. This assumption is not essential: the derivations remain valid, with the only change being an additional term involving the cross-interaction between centres and ranges.%

We now assume that the sample $y_1,\ldots,y_n$ results from projecting the set of symbolic $p$-dimensional intervals $\boldsymbol{x}_1,\ldots,\boldsymbol{x}_n$ through a fixed real-valued vector $\boldsymbol{\alpha} \in \R^p$, i.e., for each symbolic interval $y_h$ there exists a symbolic hyperrectangle $\boldsymbol{x}_h$ such that $y_h = \boldsymbol{\alpha}^T\boldsymbol{x}_h$, $h = 1,\ldots,n$.
In this new scenario, we consider the same $g$ classes, a distribution function $F_{\boldsymbol{U}}$, and the sample $\boldsymbol{x}_1 = (\boldsymbol{c}_1,\boldsymbol{r}_1, F_{\boldsymbol{U}}), \ldots, \boldsymbol{x}_n = (\boldsymbol{c}_n,\boldsymbol{r}_n, F_{\boldsymbol{U}})$. In addition, we set $\overline{\boldsymbol{x}}_B = \sum_{h=1}^n\boldsymbol{x}_h/n = (\overline{\boldsymbol{c}}, \overline{\boldsymbol{r}}, F_{\boldsymbol{U}})$ such that $\overline{y}_B = \boldsymbol{\alpha}^T\overline{\boldsymbol{x}}_B$ and $\overline{\boldsymbol{x}}_j = \sum_{h\in j}\boldsymbol{x}_h/n_j = (\overline{\boldsymbol{c}}_{j}, \overline{\boldsymbol{r}}_j, F_{\boldsymbol{U}})$, with $\overline{y}_{j} = \boldsymbol{\alpha}^T\overline{\boldsymbol{x}}_{j}$. Here, the distribution function $F_{\boldsymbol{U}}$ is only introduced to properly define the barycentres. As stated, all distribution assumptions are made on the projection space, $F_U$. 

Under this framework, we can deduce the between-scatter and within-scatter matrices in the projected space from equations \eqref{symbbetween} and \eqref{symbwithin}, whose proofs can be found in Section A of the Supplementary Materials. The between inertia is shown to be expressed by two between-scatter matrix, comparing separately the classes in the original space at the level of the centres and at the level of the ranges. That is 
\begin{equation}\label{symbbetween}
    \mathrm{BI}\ =\ %
    \sum_{j=1}^gn_j\hspace{1pt}d_M(\overline{y}_{j},\overline{y}_B)^2\ =\ \boldsymbol{\alpha}^T\boldsymbol{B}_C\boldsymbol{\alpha}\ +\ \delta |\boldsymbol{\alpha}|^T\boldsymbol{B}_R|\boldsymbol{\alpha}|,
\end{equation}
where $\boldsymbol{B}_C = \sum_{j=1}^gn_j(\overline{\boldsymbol{c}}_{j} - \overline{\boldsymbol{c}})(\overline{\boldsymbol{c}}_{j} - \overline{\boldsymbol{c}})^T$ and $\boldsymbol{B}_R = \sum_{j=1}^gn_j(\overline{\boldsymbol{r}}_{j} - \overline{\boldsymbol{r}})(\overline{\boldsymbol{r}}_{j} - \overline{\boldsymbol{r}})^T$. 

\noindent A similar conclusion is obtained for the within inertia considering the variability inside each group measured for the centres and for the ranges. Formally,
\begin{equation}\label{symbwithin}
    \mathrm{WI} \ =\ %
    \sum_{j=1}^g\sum_{h\in j} d_M(y_h,\overline{y}_{\!j})^2\ =\ \boldsymbol{\alpha}^T\boldsymbol{W}_{\!\!C}\boldsymbol{\alpha}\ +\ \delta|\boldsymbol{\alpha}|^T\boldsymbol{W}_{\!\!R}|\boldsymbol{\alpha}|,
\end{equation}
where $\boldsymbol{W}_{\!\!C} = \sum_{j=1}^g\sum_{h\in j}(\boldsymbol{c}_{h} - \overline{\boldsymbol{c}}_{j})(\boldsymbol{c}_{h} - \overline{\boldsymbol{c}}_{j})^T$ is the within-scatter matrix of the centres and $\boldsymbol{W}_{\!\!R} = \sum_{j=1}^g\sum_{h\in j}(\boldsymbol{r}_{h} - \overline{\boldsymbol{r}}_{j})(\boldsymbol{r}_{h} - \overline{\boldsymbol{r}}_{j})^T$ is the within-scatter matrix of the ranges.

We can now formulate the IFDA problem. Our goal is to find the set of orthogonal directions that maximises the ratio of between inertia and within inertia, here denoted interval Fisher ratio. We consider two orthogonality conditions: orthogonality in the usual geometric sense, and orthogonality defined with respect to the within-scatter matrix of the centres $\boldsymbol{W}_{\!\!C}$, said \textit{centre uncorrelation}, in parallel with the conventional FDA approach. The choice between two  orthogonality conditions serves to add flexibility to the model.
We introduce a matrix $\boldsymbol{M} \in \R^{p\times p}$ that regulates the orthogonality condition and the unitary norm, defined by
\begin{equation}\label{orthdef}
    \boldsymbol{M}\ \coloneqq\ \begin{cases}
    \boldsymbol{I}_p, & \mathrm{usual}\ \mathrm{orthogonality}\\[10pt]
    \displaystyle\frac{\boldsymbol{W}_{\!\!C}}{n-g}, & \mathrm{centre}\ \mathrm{uncorrelation}
    \end{cases},
\end{equation}
where $\boldsymbol{I}_p$ is the $p\times p$ identity matrix, $n$ is the sample size and $g$ is the number of classes. The matrix $\boldsymbol{M}$ is assumed to be invertible. A set of vectors $\{\boldsymbol{v}_1,\ldots,\boldsymbol{v}_r\}$, $r \in \mathbb{N}$, is said to be \textit{orthonormal} (with respect to $\boldsymbol{M}$) if $\boldsymbol{v}_j^T\boldsymbol{M}\boldsymbol{v}_k = 0$, $j \neq k$, and $\boldsymbol{v}_j^T\boldsymbol{M}\boldsymbol{v}_j = 1$, $j,k = 1,\ldots,r$.

\begin{definition}
   Interval Fisher's Discriminant Analysis is the problem of finding the set of sample orthonormal vectors in $\mathbb{R}^p$ that maximises the interval Fisher ratio, i.e.,
   \begin{equation}\label{IFDA}
    \hat{\boldsymbol{\alpha}}_i\ =\ \begin{cases}
    \underset{\boldsymbol{\alpha}:\hspace{3pt} \boldsymbol{\alpha}^T\boldsymbol{M}\boldsymbol{\alpha}\hspace{3pt} =\hspace{3pt} 1}{\mathrm{arg\ \ \  max}} \ \ \dfrac{\boldsymbol{\alpha}^T\boldsymbol{B}_C\boldsymbol{\alpha} + \delta|\boldsymbol{\alpha}|^T\boldsymbol{B}_R|\boldsymbol{\alpha}|}{\boldsymbol{\alpha}^T\boldsymbol{W}_{\!\!C}\boldsymbol{\alpha} + \delta|\boldsymbol{\alpha}|^T\boldsymbol{W}_{\!\!R}|\boldsymbol{\alpha}|}\\[15pt]
    \hat{\boldsymbol{\alpha}}_j^T\boldsymbol{M}\boldsymbol{\alpha} = 0, \ \ j \in \{1,\ldots,i-1\}
    \end{cases},\ i = 1,\ldots,s,
\end{equation}
where $s\leq p$ is the selected number of sample discriminant vectors and $\delta \in [0,1/4]$. 
\end{definition}

The sample discriminant vectors can be estimated numerically using algorithms specialised in constrained maximisation problems, as is the interval Fisher problem. We rely on the \textbf{NLopt} library \cite{NLopt}, which provides a set of algorithms that are used to solve general nonlinear minimisation problems with nonlinear constraints. Due to this, we equivalently formulate \eqref{IFDA} as the minimising the symmetric of the objective function.
In particular, we opt for the SLSQP algorithm \cite{SLSQP}, which is a local gradient-based method that requires the partial derivatives of the objective function and the equality constraint functions on the unitary norm and orthogonality. In addition, a starting value $\boldsymbol{\alpha}^{(0)}$ must be supplied. Our chosen default is a vector whose components are all equal to one. It can be the case that the algorithm converges to a local minimiser instead of a global one. In such a situation, the starting value needs to be re-evaluated or a different method combining other algorithms must be applied.
In the programming language R, the \textbf{nloptr} library serves as an interface for these procedures. The code related to the IFDA is available online in \href{https://github.com/MROS13/IFDA}{https://github.com/MROS13/IFDA}. The pseudocode related to the numerical algorithm that computes the sample discriminant vectors can be found in Algorithm~\ref{algo:ifda}.

\begin{algorithm}
\caption{IFDA Algorithm}
\label{algo:ifda}
{\bf Input:} Dataset of rectangles ($n\times p$) partitioned into $g$ classes, microdata assumption $\delta \in [0,1/4]$, orthogonality $\in$ $\{``usual", ``uncorrelated"\}$, initial vector $\boldsymbol{\alpha}^{(0)}$, number $s \leq p$ of sample discriminant vectors.\\
{\bf Output:} The ($p\times s$) sample discriminant vectors. \\
\begin{tabular}{ll}
{\footnotesize 1:} & Compute the ($p\times p$) symmetric matrices $\boldsymbol{B}_C$, $\boldsymbol{B}_{R}$, $\boldsymbol{W}_{\!\!C}$, and $\boldsymbol{W}_{\!\!R}$\\
{\footnotesize 2:} & {\bf If} orthogonality $= ``uncorrelated"$, set $\boldsymbol{M} = \boldsymbol{W}_{\!\!C}/(n-g)$; {\bf Else}, set $\boldsymbol{M} = \boldsymbol{I}_p$
\\
{\footnotesize 3:} & Define the Interval Fisher ratio $\xi(\boldsymbol{\alpha}) = \dfrac{\boldsymbol{\alpha}^T\boldsymbol{B}_C\boldsymbol{\alpha} + \delta|\boldsymbol{\alpha}|^T\boldsymbol{B}_R|\boldsymbol{\alpha}|}{\boldsymbol{\alpha}^T\boldsymbol{W}_{\!\!C}\boldsymbol{\alpha} + \delta|\boldsymbol{\alpha}|^T\boldsymbol{W}_{\!\!R}|\boldsymbol{\alpha}|}$
\\
{\footnotesize 4:} &  Set the restriction function $\sigma_0(\boldsymbol{\alpha}) = \boldsymbol{\alpha}^T\boldsymbol{M}\boldsymbol{\alpha}$ and the derivative $\sigma_0^\prime(\boldsymbol{\alpha}) = 2\boldsymbol{M}\boldsymbol{\alpha}$\\
{\footnotesize 5:} &  {\bf For} $i = 1, \ldots, s$\\
{\footnotesize 6:} & \phantom{M} Set $\sigma_{i-1} = (\sigma_0, \hat{\boldsymbol{\alpha}}_1^T\boldsymbol{M}\hat{\boldsymbol{\alpha}}_1, \ldots, \hat{\boldsymbol{\alpha}}_{i-1}^T\boldsymbol{M}\hat{\boldsymbol{\alpha}}_{i-1})$ and $\sigma_{i-1}^\prime = (\sigma_0^\prime, 2\boldsymbol{M}\hat{\boldsymbol{\alpha}}_1, \ldots, 2\boldsymbol{M}\hat{\boldsymbol{\alpha}}_{i-1})$\\
{\footnotesize 7:} & \phantom{M} Set $\hat{\boldsymbol{\alpha}}_i = nloptr(\boldsymbol{\alpha}^{(0)}, -\xi, -\xi^\prime, \sigma_{i-1}, \sigma_{i-1}^\prime, ``\textnormal{NLOPT\_LD\_SLSQP}")\$solution$ \\
{\footnotesize 8:} & \phantom{M} {\bf If} $\hat{\boldsymbol{\alpha}}_i$ is the zero vector or undefined, set $s=i-1$ and {\bf Break}\\
{\footnotesize 9:} & \phantom{M} Update $\hat{\boldsymbol{\alpha}}_i = \dfrac{\hat{\boldsymbol{\alpha}}_i}{\sqrt{\hat{\boldsymbol{\alpha}}_i^T\boldsymbol{M}\hat{\boldsymbol{\alpha}}_i}}$\\
{\footnotesize 10:} & {\bf End}\\
{\footnotesize 11:} & {\bf Return} $\hat{\boldsymbol{\alpha}}_1, \ldots, \hat{\boldsymbol{\alpha}}_s$
\end{tabular}
\end{algorithm}

The computation of the derivatives is fairly straightforward, with the exception of the objective function and the absolute value vector which is not usually encountered in the literature. In the next proposition, we present a list of partial derivatives of functions of the absolute value vector.

\begin{proposition}\label{lemmaderivatives}
Let $\boldsymbol{\alpha} = (\alpha_1,\ldots,\alpha_p)^T\in \mathbb{R}^p$, $p\in\mathbb{N}$, where $\alpha_i \neq 0$, $i=1,\ldots,p$. In addition, let $\boldsymbol{b}\in \mathbb{R}^{p}$ be a constant vector and $\boldsymbol{B} \in \mathbb{R}^{p\times p}$ be a constant matrix that does not depend on $\boldsymbol{\alpha}$. The following assertions hold:
\begin{itemize}
    \item[(1)]\hspace{5pt}\label{Eq:Prop3.5.1}   $\dfrac{\partial (\boldsymbol{b}^T|\boldsymbol{\alpha}|)}{\partial\boldsymbol{\alpha}}\ =\ \mathrm{diag}(\mathrm{sgn}(\boldsymbol{\alpha}))\boldsymbol{b}\hspace{1pt}.$
    \item[(2)] \hspace{5pt}\label{Eq:Prop3.5.2} $\dfrac{\partial (|\boldsymbol{\alpha}|^T\boldsymbol{B}|\boldsymbol{\alpha}|)}{\partial\boldsymbol{\alpha}}\ =\ \mathrm{diag}\big(\mathrm{sgn}(\boldsymbol{\alpha})\big)(\boldsymbol{B} + \boldsymbol{B}^T)|\boldsymbol{\alpha}|$
    \item[] \hspace*{94.5pt} $=\ 2\hspace{2pt}\mathrm{diag}\big(\mathrm{sgn}(\boldsymbol{\alpha})\big)\boldsymbol{B}|\boldsymbol{\alpha}|,$\quad if $\boldsymbol{B}$ is symmetric\hspace{2pt}$.$
    \item[(3)] \label{Eq:Prop3.5.3}  Let $\fun(\boldsymbol{\alpha})\ \coloneqq\   \dfrac{\gamma(\boldsymbol{\alpha})}{\beta(\boldsymbol{\alpha})}\ =\ \dfrac{\boldsymbol{\alpha}^T\boldsymbol{B}_C\boldsymbol{\alpha} + \delta|\boldsymbol{\alpha}|^T\boldsymbol{B}_R|\boldsymbol{\alpha}|}{\boldsymbol{\alpha}^T\boldsymbol{W}_{\!\!C}\boldsymbol{\alpha} + \delta|\boldsymbol{\alpha}|^T\boldsymbol{W}_{\!\!R}|\boldsymbol{\alpha}|}$ be the interval Fisher ratio. Then,
    \item[] $\fun^\prime(\boldsymbol{\alpha})\ \coloneqq\ \dfrac{\partial \fun(\boldsymbol{\alpha})}{\partial \boldsymbol{\alpha}}\ =\ \dfrac{2}{\beta(\boldsymbol{\alpha})}\Bigg[\big(\boldsymbol{B}_C - \boldsymbol{W}_{\!\!C} \hspace{1pt}\fun(\boldsymbol{\alpha}) \big)\boldsymbol{\alpha} + \delta \mathrm{diag}(\mathrm{sgn}(\boldsymbol{\alpha})) \big(\boldsymbol{B}_R - \boldsymbol{W}_{\!\!R} \hspace{1pt}\fun(\boldsymbol{\alpha})\big)|\boldsymbol{\alpha}| \Bigg]\hspace{1pt}.$
\end{itemize} 
\end{proposition}

\begin{proof}
    See Section B of the Supplementary Materials. 
\end{proof}

Let $\hat{\boldsymbol{\alpha}}_1,\ldots,\hat{\boldsymbol{\alpha}}_s$ be the sample discriminant vectors retained. In IFDA, a new observation $\boldsymbol{x}_0$ is assigned to the class that minimises the squared Mallows' distance between the projected observation and the projected class barycentres, in the space spanned by the discriminant vectors. In more detail, let \mbox{$\boldsymbol{x}_0 = (\boldsymbol{c}_0,\boldsymbol{r}_0, F_{\boldsymbol{U}})$} and construct the symbolic hyperrectangle $\hat{\boldsymbol{y}}_0 = (\hat{y}_1,\ldots, \hat{y}_s)^T$ from the sample discriminants $\hat{y}_t = (\hat{\boldsymbol{\alpha}}_t^T\boldsymbol{c}_0,|\hat{\boldsymbol{\alpha}}_t|^T\boldsymbol{r}_0, F_U)$, $t= 1,\ldots,s$, and $\overline{\boldsymbol{y}}_j = (\overline{y}_{j1}, \ldots, \overline{y}_{js})^T$ from the projected sample class barycentres $\overline{y}_{jt} = \hat{\boldsymbol{\alpha}}_t^T\overline{\boldsymbol{x}}_j = (\hat{\boldsymbol{\alpha}}_t^T\overline{\boldsymbol{c}}_j, |\hat{\boldsymbol{\alpha}}_t|^T\overline{\boldsymbol{r}}_j, F_U)$, $j = 1,\ldots,g$, following Moore's algebraic structure \eqref{moore}. 
 It is assumed that the symmetric latent distribution $F_U$ is common to every projected symbolic interval such that \eqref{Eq:mallowsdelta} holds.
The decision rule is to assign $\boldsymbol{x}_0$ to class $\hat{\jmath} \in \{1,\ldots,g\}$, whenever 
\begin{align}\label{IFDAclassify}
     \hat{\jmath}\ &=\ \underset{j \hspace{1pt} \in \hspace{1pt}\{1,\ldots,g\}}{\mathrm{arg\ \ \ min}} \ \ d_M(\hat{\boldsymbol{y}}_0, \overline{\boldsymbol{y}}_j)^2\notag\\
     &=\ \underset{j \hspace{1pt} \in \hspace{1pt}\{1,\ldots,g\}}{\mathrm{arg\ \ \ min}} \ \ \sum_{t=1}^s \left(\hat{\boldsymbol{\alpha}}_t^T\boldsymbol{c}_0 - \hat{\boldsymbol{\alpha}}_t^T\overline{\boldsymbol{c}}_j\right)^2 +\hspace{2pt} \delta\left(|\hat{\boldsymbol{\alpha}}_t|^T\boldsymbol{r}_0 - |\hat{\boldsymbol{\alpha}}_t|^T\overline{\boldsymbol{r}}_j\right)^2.
\end{align}

\section{Visualising Classification Results}\label{sec:Visualization}

The performance of a classifier can be assessed using both class-specific and global measures. The most common class-specific measures are the recall of the $j$-th class, $\Rec(j)$, defined as the probability that an observation from class $j$ is correctly classified, and the precision of the $j$-th class, $\Pr(j)$, defined as the probability that an observation assigned to class $j$ truly belongs to it. A widely used global measure is the accuracy (Acc). Theoretically, accuracy is defined as the probability that a classifier assigns the correct label, while, in practice, it is estimated from classification results on a test set with known labels, as is the case with the other measures. These estimators assume that the class labels in both training and test sets are error-free. Nevertheless, such measures neither reflect the classifier’s confidence in its assignments nor capture the influence of outliers. Graphical displays of classifier performance can complement them by providing additional insight and enhancing interpretability. 

In \cite{RaymaekersSih:2022,RaymaekersClassmap:2022}, the authors introduced graphical visualisation tools to improve the interpretability of classification results. In this work, we extend these tools to the interval-valued data setting, providing a new contribution to the symbolic framework.

The stacked mosaic plot visualises both the estimated prior probabilities $p_j$ of the classes and  recall values $\Rec(\hat{\jmath},j)$, where $\Rec(\hat{\jmath},j)$ denotes the estimated proportion of observations from the $j$-th true class that the classifier assigns to the $\hat{\jmath}$-th class (with $\Rec(j,j)=\Rec(j)$). For simplicity, the same notation is used for the theoretical measures and their estimates. 
The plot is a unit square divided into $g$ vertical sections along the x-axis, each section representing one true class having width $p_j$. Inside each section, rectangles are stacked vertically to reflect the recalls: the rectangle for $\Rec(j)$ appears at the bottom, corresponding to correctly classified observations, whilst the others, of heights $\Rec(\hat{\jmath},j)$ for $\hat{\jmath} \neq j$, correspond to misclassifications in one of the remaining classes. 
This construction makes the plot directly comparable to a normalised confusion matrix, with the normalization carried out conditionally on the true class. As a result, the plot provides an integrated view of classification performance across all classes.

To illustrate the stacked mosaic plot, consider the \emph{credit card} dataset \cite{BillardDiday2003,BillardDiday2006,Oliveira2021}, which contains the monthly expenses of three credit card users measured on five interval-valued variables over the course of one year. To exemplify, we focus on two of these variables: Food ($x_1$) and Gas ($x_2$). This results in a dataset of $n = 36$ observations on $p = 2$ variables. The confusion matrix resulting from the interval Fisher classifier is shown in Figure \ref{mat:without_outlier}, where we used all observations to train and test. The estimated accuracy is $0.861$. 
\begin{figure}[htbp]
\centering
\begin{subfigure}{0.40\textwidth}
\centering
\[
\left( \begin{array}{ccc}
11& 0& 1\\
2& 8& 2\\
0& 0& 12\\
\end{array} \right)
\]
\caption{Without global outlier class.}
\label{mat:without_outlier}
\end{subfigure}
\hfill
\begin{subfigure}{0.55\textwidth}
\centering
\[
\left( \begin{array}{cccc}
10& 0& 1&1\\
2& 8& 2&0\\
0& 0& 12&0\\
\end{array} \right)
\]
\caption{With the class of global outliers (last column).}
\label{mat:with_outlier}
\end{subfigure}
\caption{Comparison of confusion matrices for the \emph{credit card} dataset. The rows correspond to the true classes and the columns to the predicted classes.}
\label{fig:confusion_matrices}
\end{figure}

\begin{figure}[htbp]
\centering
\begin{subfigure}{0.48\textwidth}
    \centering
    \includegraphics[width=\linewidth]{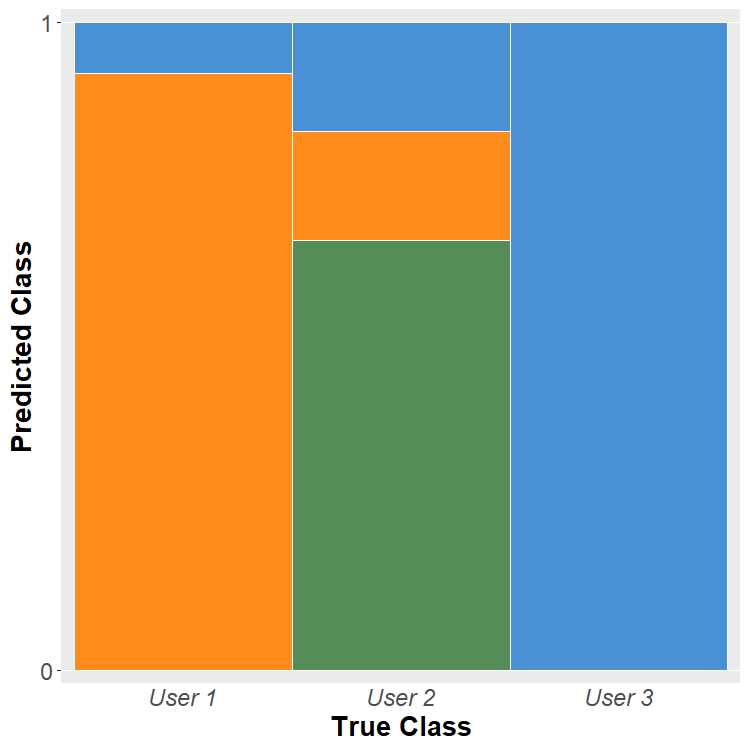}
    \caption{Without outlier detection.}
    \label{fig:Mosai_credit_alldataset}
\end{subfigure}
\hfill
\begin{subfigure}{0.48\textwidth}
    \centering
    \includegraphics[width=\linewidth]{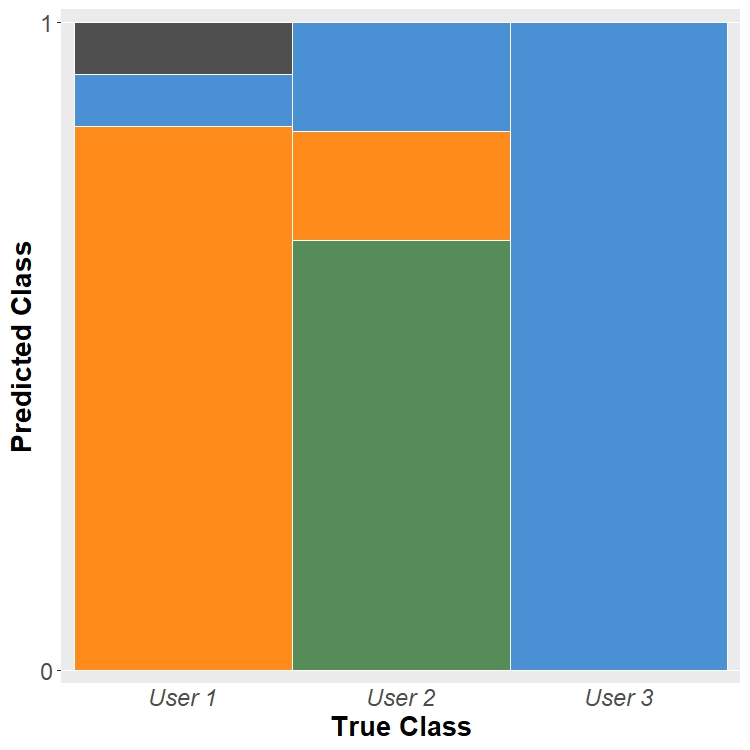}
    \caption{With outlier detection ($\tau=0.95$).}
    \label{fig:Mosai_credit_alldataset_out}
\end{subfigure}
\caption{Stacked mosaic plots for the \emph{credit card} dataset with variables Food and Gas. The dark grey rectangle represents the proportion of global outliers in class 1.}
\label{fig:Mosaic_credit}
\end{figure}

Since for each user we observe 12 months, the estimated prior probabilities are $p_j = 1/3$ for $j = 1,2,3$. According to the confusion matrix, the classifier incorrectly assigns one of User 1’s monthly expenses to User 3. Consequently, $\Rec(3,1) = 1/12$ and $\Rec(1) = 11/12$. Therefore, in Figure \ref{fig:Mosai_credit_alldataset}, the height of the first orange rectangle (representing User 1) is $11/12$, while the second, shown in blue, has height $1/12$. 
Figure \ref{fig:Mosai_credit_alldataset} shows that the classifier correctly classifies all of User 3’s expenses, while most errors occur for User 2. Specifically, the classifier confuses User 2 equally often with User 1 and with User 3. This is illustrated by the orange (User 1) and blue (User 3) rectangles of equal height stacked on the second bar, which represents User 2 (with first rectangle in green). 

Given the squared Mallows' distance of an observation to each class barycentre in the projected space, we apply transformations such as the robust Yeo–Johnson transformation (see \cite{Raymaekers.Rousseuw:2024} for details) to obtain values that approximately follow a standard normal distribution. For the $j$-th class, this enables the estimation of the probability that a distance in that class is less than or equal to that of the $h$-th observation. Large distances indicate that the observation is atypical for that class. We refer to this estimated probability as \textit{farness}, and denote the farness of the $h$-th observation with respect to class $j$ by ${\rm farness}(h,j)$, for $h = 1,\ldots,n$ and $j = 1,\ldots,g$.
If the true class of the $h$-th observation is $j_h$, we define ${\rm farness}(h) \coloneqq {\rm farness}(h,j_h)$.

For a given threshold $\tau\in [0,1]$, the $h$-th observation from class $j_h$ is considered a \textit{local outlier} if ${\rm farness}(h) > \tau$. 
The choice of $\tau$ reflects the approximate probability $\tau$ of incorrectly labelling a regular observation of that class as outlier. 

An observation that is assigned as atypical for all classes is seen as a \textit{global outlier}, and a new outlier detection rule can be defined as
\begin{equation}
\label{eq:Ofarness.Out.rule}
    {\rm If} \; {O}(h):= \min_{j=1,\ldots , g} {\rm farness}(h,j) \ > \ \tau , {\rm \; the \;}  h{\rm\textnormal{-}th \; observation } {\; \rm is\; a\;  global\;  outlier. \;} 
\end{equation}

In the stacked mosaic plot, for each class, the corrected classified observations can be divided into two subclasses: global outliers, verifying ${O}(h) > \tau$ and regular, otherwise. Then, the original $\Rec(j)$ is divided into two, being the proportion of global outliers represented in the stacked mosaic plot by a dark grey rectangle.  

In the credit card example, with $\tau=0.95$, the July expense ($h=7$) of User 1 is correctly attributed to this user. However, due to its high value of farness in all classes ($O(7)=0.979 > 0.95$), the observation is classified as a global outlier. Consequently, the new stacked mosaic plot is based on the updated confusion matrix shown in Figure \ref{mat:with_outlier}, where the fourth column corresponds to the global outlier class. Accordingly, in Figure \ref{fig:Mosai_credit_alldataset_out}, the dark grey rectangle has height $1/12$. 

\begin{figure}[htbp]
\centering
    \includegraphics[width = 0.8\linewidth]{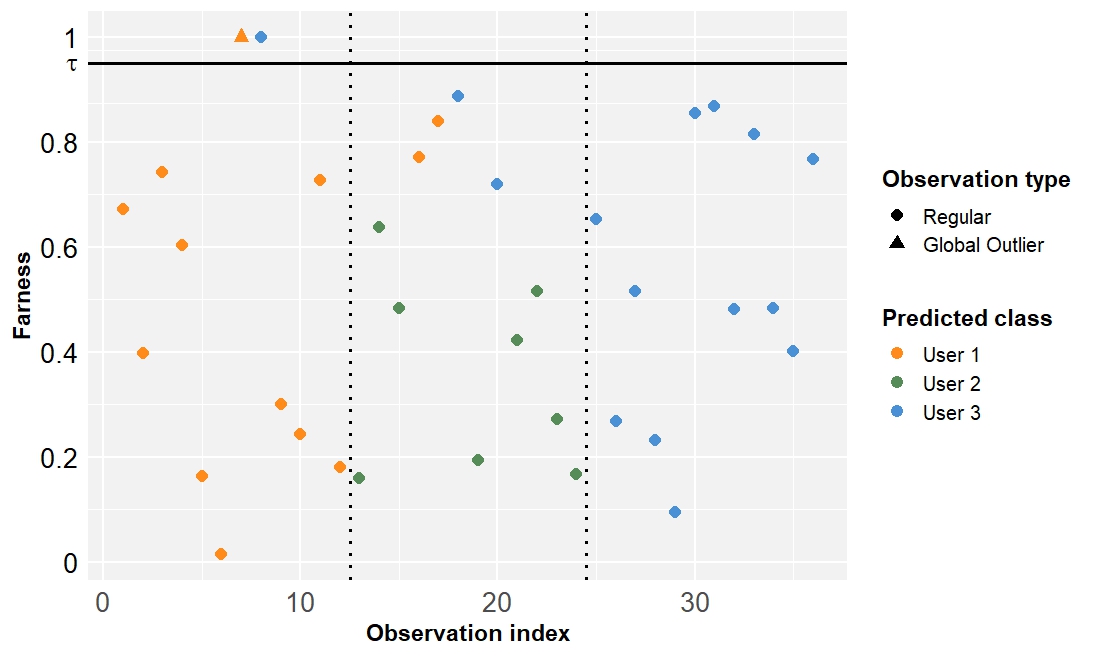}
    \caption{Farness plot for the credit card dataset. Vertical dotted lines separate the indices of the true classes. The horizontal line makes the threshold $\tau = 0.95$.}
    \label{fig:Farness_credit_alldataset}
\end{figure}

The farness of each observation can be visualised similarly to outlier detection based on the robust Mahalanobis distance \cite{FastMCD:1999}. In Figure \ref{fig:Farness_credit_alldataset}, observations are colored by their predicted class, while triangles indicate observations with $O(h) > \tau$ (in this example $\tau = 0.95$), classified as global outliers. Observation's indices are ordered by true class, with vertical lines marked the index class separation. 

Given an observation, it is important to assess whether the classifier assigns it to the correct class or to a different one. A misclassification may result from the classifier’s limited ability to discriminate between classes (the traditional interpretation) or because the observation shares strong similarities with the predicted class. In the latter case, a labelling error may explain the apparent misclassification.

Inspired by the Probability of Alternative Class index proposed by \cite{RaymaekersSih:2022,RaymaekersClassmap:2022}, we introduce a measure that compares the distance of a point to its assigned class with the distance to its closest alternative class.
Let $(\hat{\bm{y}}_h, j_h)$ denote the pair consisting of the explanatory variables projected on the subspace induced by IFDA, $\hat{\bm{y}}_h$, and the true class label of the $h$-th observation, $j_h$. Define $d_M(h)\coloneqq d_M(\hat{\bm{y}}_h,\overline{\bm{y}}_{j_h})$ as the Mallows' distance between $\bm{y}_h$ and the sample barycentre of class $j_h$ in the same projected subspace, $\overline{\bm{y}}_{j_h}$.
The distance to the best alternative class (DAC) is defined as
\begin{equation}
\label{eq:BestAlternativeClass}
{\rm DAC}(h) = \min_{j \neq j_h} d_M(\hat{\bm{y}}_h,\overline{\bm{y}}_{j}).
\end{equation}
To compare $d_M(h)$ with ${\rm DAC}(h)$, we introduce a new index, the logistic transformation of the deviation to the best alternative class (${\rm \ell DAC}$), as the logistic function of the difference between the two distances, $d_M(h)-{\rm DAC}(h)$:
\begin{equation}
    \label{eq:DPC}
    {\rm \ell DAC}(h):=\frac{1}{1+e^{-(d_M(h)-\rm{DAC}(h))}}.
\end{equation}

This transformation changes the scale to $[0,1]$ making the interpretation clearer. If $d_M(h) \ll \rm{DAC}(h)$, then $\hat{\bm{y}}_h$ lies much closer to the barycentre of its true class than to that of the best alternative, which implies a correct classification and ${\rm \ell DAC}(h) \ll 1/2$.
Conversely, if $d_M(h) \gg \rm{DAC}(h)$, the observation resembles the alternative class more closely, leading to misclassification and ${\rm \ell DAC}(h) \gg 1/2$.
When $\rm{DAC}(h)$ $\approx d_M(h)$, the distances to the true and alternative class are similar, making the assignment uncertain, with ${\rm \ell DAC}(h) \approx 1/2$. 
Note that if ${\rm \ell DAC}(h) \approx 0$ (respectively, ${\rm \ell DAC}(h) \approx 1$),  the classifier strongly favours assignment to the true (respectively, wrong) class.

\begin{figure}[htbp]
\centering
\begin{subfigure}{0.49\textwidth}
    \centering
    \includegraphics[width=\linewidth]{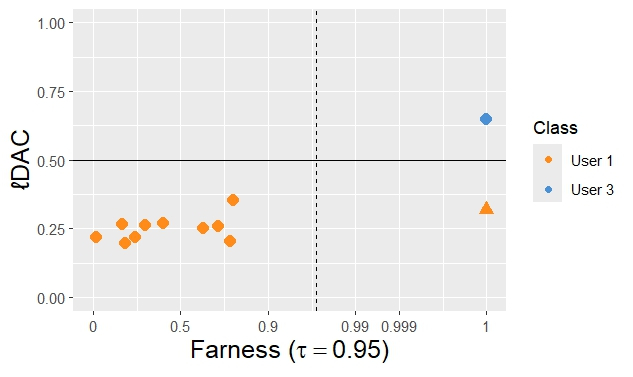}
    \caption{User 1.}
    \label{fig:DAC_credit_alldataset_1}
\end{subfigure}
\hfill
\begin{subfigure}{0.49\textwidth}
    \centering
    \includegraphics[width=\linewidth]{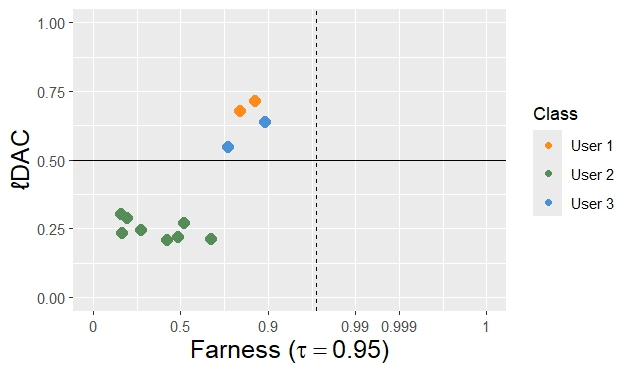}
    \caption{User 2.}
    \label{fig:DAC_credit_alldataset_2}
\end{subfigure}
\caption{Class map plots for the first two users of the \emph{credit card} dataset considering Food, Gas, and the entire dataset ($\tau=0.95$).}
\label{fig:DAC_credit}
\end{figure}

By plotting farness versus ${\rm \ell DAC}$ for each observation, we obtain an integrated view and interpretation of the test set and the classifier’s behaviour. Observations with ${\rm \ell DAC}$ values below 0.5 and farness below $\tau$ are interpreted as regular points correctly assigned to their true class. In Figure \ref{fig:DAC_credit}, these observations are highlighted in the bottom left section of the plot.

In Figure \ref{fig:DAC_credit_alldataset_1}, the blue point in the top right section represents an observation from User 1 misclassified as User 3. Its high farness relative to User 1 indicates that the classifier considers it far from its true class but a reasonable fit to User 3. The orange triangle in the bottom right section corresponds to a correctly assigned User 1's monthly expense that is far from all class barycentres, identifying it as a global outlier. 

In Figure \ref{fig:DAC_credit_alldataset_2}, four observations are misclassified (top left): two of User 1 and two of User 3. The point with ${\rm \ell DAC}$ closest to 0.5 amongst these observations is not far from its true class compared with the distance to the assigned class.

\begin{figure}
    \centering
    \includegraphics[width=0.8\linewidth]{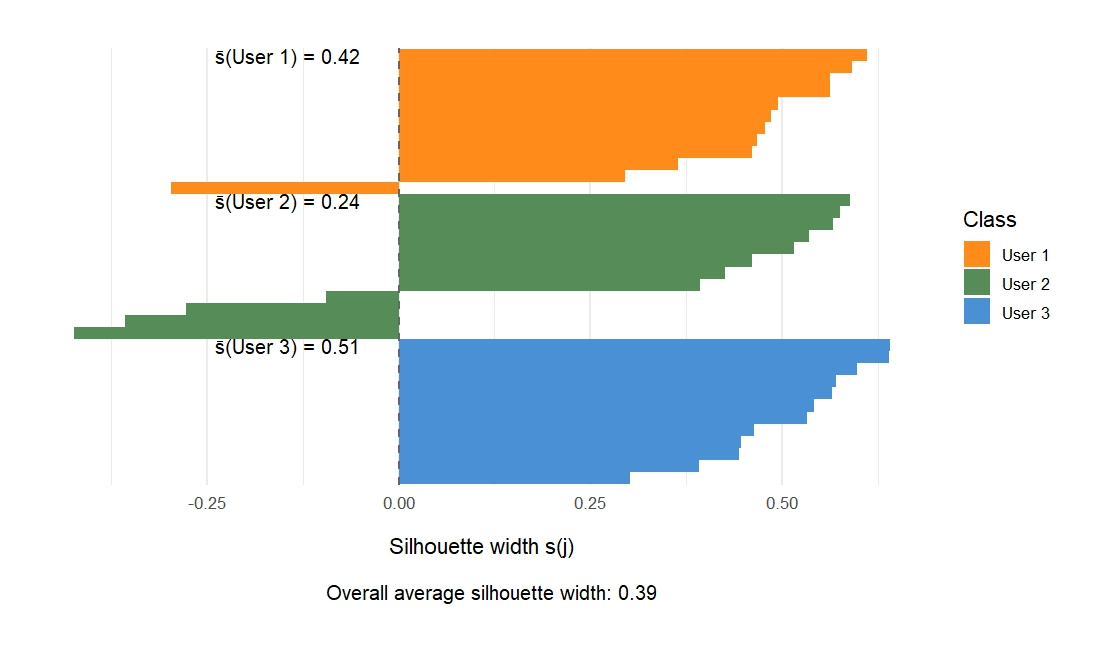}
    \caption{Silhouette plot for the credit card dataset using the IFDA classifier.}
    \label{fig:Silhplot}
\end{figure}

Following the silhouette plot proposed for supervised problems in \cite{RaymaekersSih:2022} and given the interpretation of ${\rm \ell DAC}$, we adopt the definition $s(j):=1-2{\rm \ell DAC}(j)$. This coefficient ranges between $-1$ and $1$, with an interpretation analogous to the original silhouette index introduced for cluster analysis in \cite{Rousseeuw1987}. Strongly positive values indicate that the classifier fits the observation well to its true class, while strongly negative values suggest a better fit to an incorrect class. 
In Figure \ref{fig:Silhplot}, we present the silhouette plot for the credit card dataset. The average silhouette values by user are $\bar{s}({\rm User}\ 1) = 0.42$, $\bar{s}({\rm User}\ 2) = 0.24$, and $\bar{s}({\rm User}\ 3) = 0.51$, respectively, with an overall average of $0.39$. The figure highlights the strong predictive performance for User 3, which attains the highest average silhouette (0.51), while User 2 shows comparatively weaker performance, with the lowest average silhouette (0.24), indicating a higher proportion of misclassified observations in this class.

\section{Numerical studies}\label{sec:NumStudies}

In this section, we examine two practical examples in which we apply the proposed methodology to classify interval-valued data. The simulation study focuses on analysing a variety of scenarios under
different conditions to understand the strengths and weaknesses of the symbolic method. In addition, we present a real example regarding the detection of redirected traffic to demonstrate the applicability of IFDA and the visualisation tools discussed.

\subsection{Simulation Study}\label{sec:SimStudy}

This simulation study evaluates the classifier’s performance for two classes that may exhibit small or large separation between classes in their centres, ranges, or both. The simulation design is inspired by the scenarios presented in \cite{Dias2021}. 
We consider three symbolic intervals ($p = 3$) and two classes ($g = 2$). The centres and ranges for each class and variable are generated from continuous uniform distributions, with parameters depending on the simulation scenario. For class 1, and in all scenarios, the expected values of the centres and ranges are fixed as follows: $c_1 = 20$, $r_1 = 16$; $c_2 = 10$, $r_2 = 12$; $c_3 = 5$, and $r_3 = 8$. For $i = 1,2,3$, the centres of the $i$-th variable are generated by $\mathrm{Unif}[0.6c_i(1+a), 1.4c_i(1+a)]$, and the ranges by $\mathrm{Unif}[0.6r_i(1+b), 1.4r_i(1+b)]$, where $a = b = 0$ for the class $1$.
For class $2$, four simulation cases are considered to characterise the degree of separation between classes in terms of centres and ranges:
\textbf{Case A}: Large variation in both centres and ranges between classes: $a = 0.6$, $b = 0.6$. 
\textbf{Case B}: Small variation in both centres and ranges between classes: $a = 0.2$, $b = 0.2$; 
\textbf{Case C}: Small variation in centres and minimal variation in ranges between classes: $a = 0.2$, $b = 0.05$; 
\textbf{Case D}: Minimal variation in centres and small variation in ranges: $a = 0.05$, $b = 0.2$; 

We consider two configurations for the training sets: equal sizes with $n_1 = n_2 = 125$ (identified by $p_1=0.5$, the estimated prior probability of the first class), and unequal sizes with $n_1 = 50$ and $n_2 = 200$ ($p_1=0.2$). The corresponding test sets are five times larger than their respective training sets.
Each simulation scenario is replicated $m = 100$ times, with $\delta$ fixed at $1/12$.

The classifier’s performance is evaluated using the macro $\mathrm{F}_1$ measure, defined as the mean of the $\mathrm{F}_1$ measures across the two classes, and the G-mean, defined as the geometric mean of the two classes’ recalls. For the $j$th class, the recall is the proportion of correctly assigned observations, while the precision is the proportion of observations assigned to class $j$ that truly belong to that class. Finally, the $\mathrm{F}_1$ measure for a given class is the harmonic mean of its precision and recall. 

\begin{figure}[h!] \centering \includegraphics[width=\linewidth]{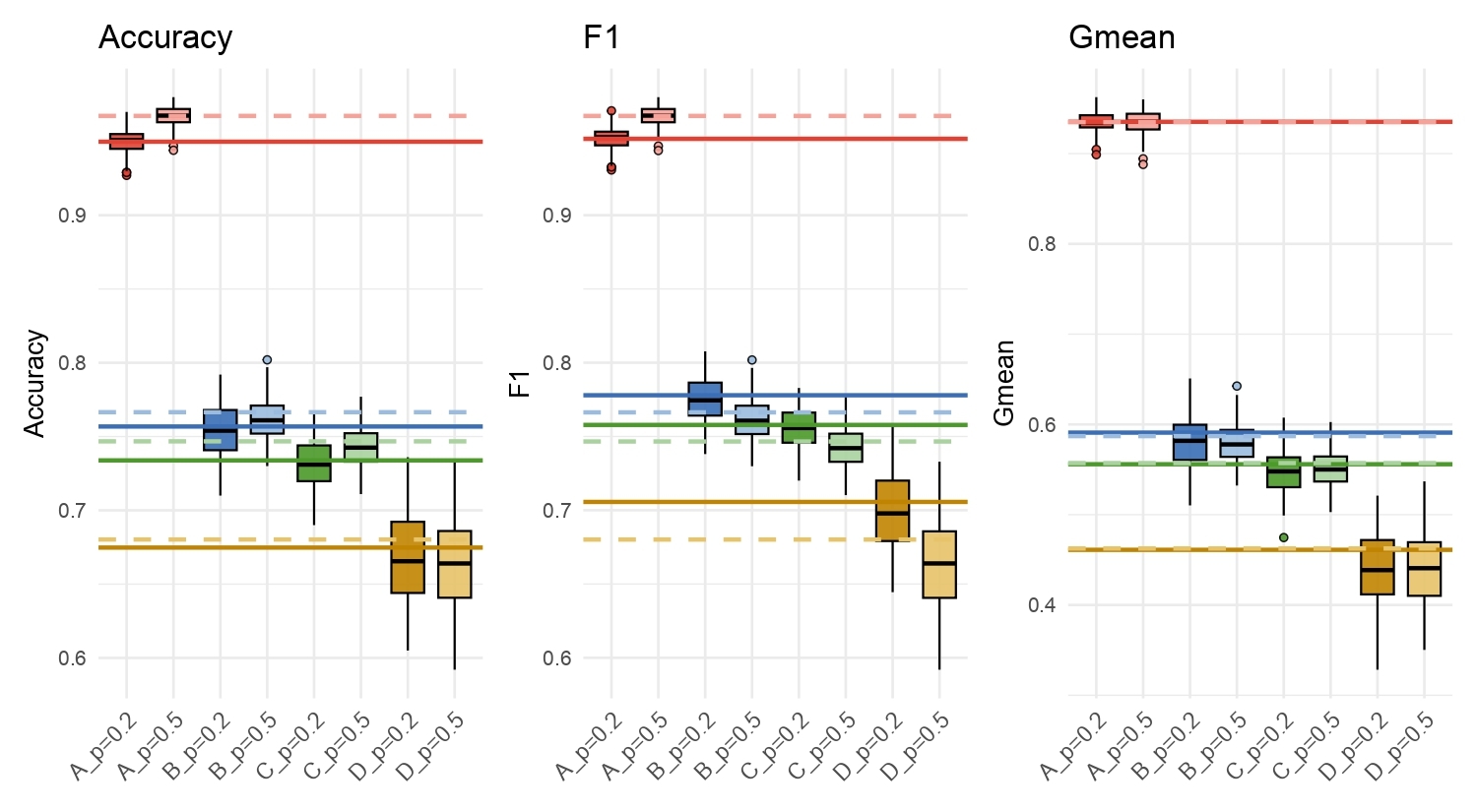}
    \caption{Performance measures of the different simulation scenarios, using one sample discriminant vector. The continuous (respectively, dashed) line represents the mean theoretical value for the unequal $p_1=0.2$ (equal, $p_1=0.5$) classes' sample size. }
    \label{fig:simulRes}
\end{figure}

To benchmark our approach, we computed the theoretical within- and between-class covariance matrices for each simulation scenario and evaluated performance on the test sets. The sample mean served as the reference. Figure \ref{fig:simulRes} summarises the results: colors distinguish Cases A–D, with darker shades corresponding to unequal class sizes ($p_1=0.2$) and lighter shades to equal sizes ($p_1=0.5$). Solid and dashed lines denote the respective benchmark references.

The best performance is obtained when the classes are clearly separated (Case A). When both centres and ranges differ only slightly (Case B), performance decreases, reflecting the reduced class separation, while separation limited to the centres produces a minor additional decline. The poorest results occur when separation is small and restricted to the ranges. In this case, the proposed method underestimates all performance measures, as the contribution of the ranges to the Mallows' distance is downweighted by $\delta = 1/12$, diminishing their effect when they are the sole source of class distinction. 
Theoretical results indicate that, on average, accuracy is lower for the unbalanced case ($p_1=0.2$) than for the balanced case ($p_1=0.5$), as expected. Except for Case A, the opposite occurs for the $\mathrm{F}_1$ measure, reflecting the influence of precision on this metric. Since the G-mean depends solely on class recall, its reference values are similar for both balanced and unbalanced scenarios. The results obtained with our method follow the same overall pattern.

To complete the comparison, we computed the Mean Squared Error (MSE) between the estimated and theoretical performance measures. The results, shown in Table \ref{tab:mse_acv}, confirm that lower class separation yields higher MSE, while clearer separation reduces it.

To assess the proximity between the theoretical discriminant vector $\boldsymbol{\alpha}_1$ and its estimate $\hat{\boldsymbol{\alpha}}_1$, we used one minus the Absolute Cosine Value (ACV), where 

\begin{equation}
    {\rm{ACV}}(\hat{\boldsymbol{\alpha}_1}) = \frac{1}{m}\sum_{j=1}^m \frac{|\hat{\boldsymbol{\alpha}}_{1j}^T\boldsymbol{\alpha}_1|}{||\hat{\boldsymbol{\alpha}}_{1j}||\ ||\boldsymbol{\alpha}_1|| },
\end{equation}

\noindent and $\hat{\boldsymbol{\alpha}}_{1j}$ denotes the $j$th estimate of $\boldsymbol{\alpha}_1$. The vector $\boldsymbol{\alpha}_1$ is computed by feeding the theoretical between- and within-scatter matrices to our classifier. 
The 1-ACV coefficient ranges from 0 to 1, where smaller values indicate greater proximity between the estimated and true vectors.

Table \ref{tab:mse_acv} shows that cases with greater class separation produce more accurate estimates of $\boldsymbol{\alpha}_1$, and balanced class proportions further improve performance.

\begin{table}[ht]
\centering
\caption{Mean Squared Error (MSE) and $1-\mathrm{ACV}$ ($1$ $-$ Absolute Cosine Value) values for Cases A–D under different sample proportions ($p_1=0.2$ and $p_1=0.5$).}
\label{tab:mse_acv}
\begin{tabular}{lccccc}
\hline
 &&  \multicolumn{4}{c}{\textbf{Cases}}\\\cline{3-6}
\textbf{Measure} & $p_1$ & \textbf{A} & \textbf{B} & \textbf{C} & \textbf{D} \\
\hline
\multirow{2}{*}{MSE(Acc)} & $0.2$   & 0.0010 & 0.0055 & 0.0065 & 0.0198 \\
 & $0.5$   & 0.0004 & 0.0021 & 0.0019 & 0.0266 \\\hline
\multirow{2}{*}{MSE($\mathrm{F}_1$)} & $0.2$ & 0.0009 & 0.0038 & 0.0045 & 0.0154 \\
 & $0.5$ & 0.0004 & 0.0021 & 0.0019 & 0.0268 \\\hline
\multirow{2}{*}{MSE(G-mean)} & $0.2$ & 0.0014 & 0.0086 & 0.0064 & 0.0553 \\
 & $0.5$ & 0.0014 & 0.0051 & 0.0043 & 0.0470 \\\hline
\multirow{2}{*}{$1-\mathrm{ACV}(\hat{\boldsymbol{\alpha}}_1)$} & $0.2$ & 0.0023 & 0.0088 & 0.0107 & 0.1126 \\
& $0.5$ & 0.0022 & 0.0061 & 0.0072 & 0.0569 \\
\hline
\end{tabular}
\end{table}

\subsection{Real Example: Traffic Redirection Attacks}\label{sec:RealExample}
In \cite{salvador2014rtt}, the problem of detecting traffic redirection attacks was studied using monitoring probes deployed across multiple geographic regions. Each probe measured round-trip times (RTTs) for groups of 10 packets, retaining the minimum and maximum values as the macrodata. Increases in RTT were interpreted as evidence of packet hijacking by an intermediate relay, i.e., a man-in-the-middle attack. A comprehensive analysis of these datasets is provided in \cite{Subtil.et.al:2023}.

We analyse two datasets targeting Hong Kong (HK) and London (LN), representing the locations of the intended target machines. Measurements were collected from $p=12$ probes located in Amsterdam, Frankfurt2, Milan, Stockholm, Hafnarfjordur (Iceland), Petah Tikva (Israel), Chicago2, Los Angeles2, Vi\~na del Mar (Chile), São Paulo2, Johannesburg1, and Johannesburg2. Traffic redirection was routed through four relays: Los Angeles1, Madrid, Moscow, and São Paulo. The Hong Kong dataset contains $n_{\rm HK}=11496$ observations, and the London dataset contains $n_{\rm LN}=11434$. Each observation is assigned to one of five classes: regular traffic (no redirection) or one of the four relays corresponding to an attack.

Because the microdata are unavailable, the variables $U_i$ are assumed to be symmetric and identically distributed. The parameter $\delta$ and the number of discriminant vectors $s$ are selected via a grid search ($\delta \in \{0, 0.01, \ldots, 0.25\}$ and $s \in \{2, \ldots, 9\}$), retaining the combination that maximizes classification accuracy. The first 60\% of the observations are used for training (including the estimation of $\delta$ and $r$), and the remaining 40\% are reserved for testing. During training, the data are repeatedly divided into two equal, stratified subsets across 30 random splits. The pair ($\delta, r$) yielding the highest mean accuracy is then used to re-estimate the classifier on the full training set. 

\begin{figure}[htbp]
\centering
\begin{subfigure}{0.48\textwidth}
    \centering
    \includegraphics[width=\linewidth]{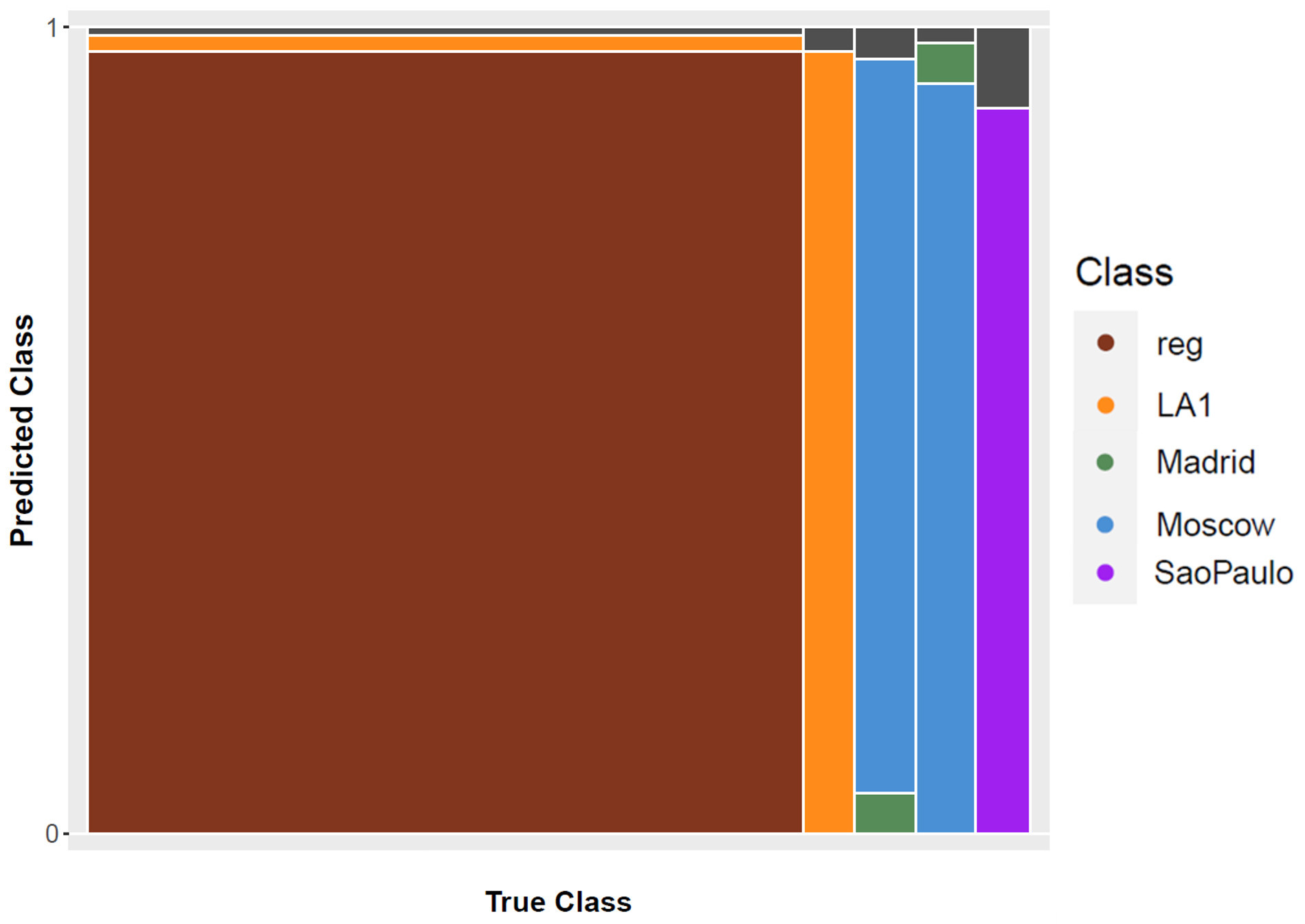}
    \caption{Hong Kong.}
    \label{fig:Mosaic_HK}
\end{subfigure}
\hfill
\begin{subfigure}{0.48\textwidth}
    \centering
    \includegraphics[width=\linewidth]{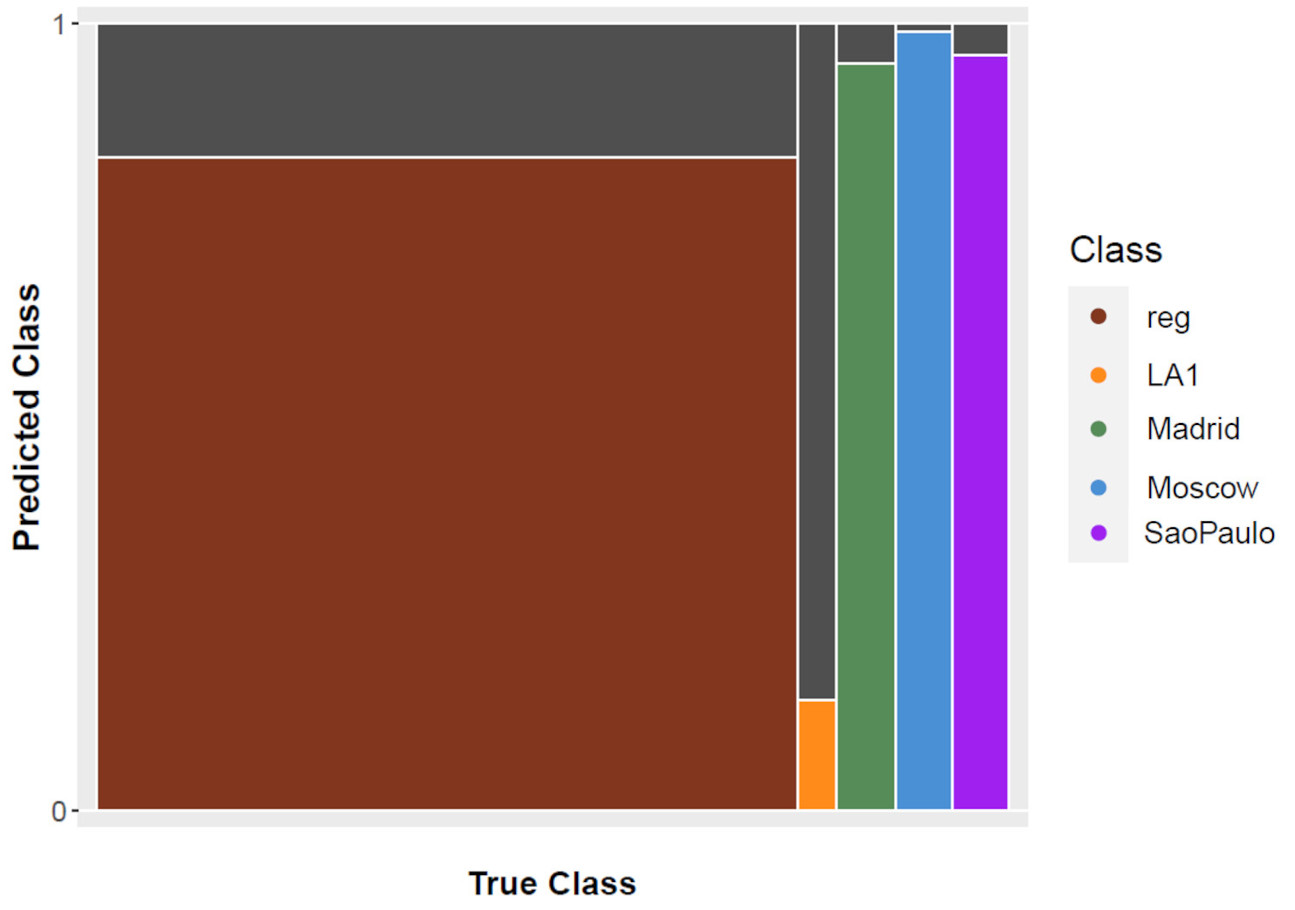}
    \caption{London.}
    \label{fig:Mosaic_London}
\end{subfigure}
\caption{Stacked mosaic plots for the RTT datasets ($\tau=0.99$).}
\label{fig:Mosaic_RTT}
\end{figure}

For the Hong Kong dataset, we obtained $\delta_{\rm HK}=0.19$ with $r_{\rm HK}=8$, and $(\delta_{\rm LN}, r_{\rm LN})=(0.01, 2)$ for the London dataset, yielding accuracies of $0.917$ and $1.000$, respectively. The corresponding confusion matrices are presented in Figure \ref{fig:confusion_matrices_RTT}. Under $\tau=0.99$, the stacked mosaic plots in Figure~\ref{fig:Mosaic_RTT} illustrate the classification results and highlight the detection of global outliers. 

\begin{figure}[htbp]
\centering
\begin{subfigure}{0.45\textwidth}
\centering
\[
\left( \begin{array}{ccccc}
3396&   84&      0&      1&        0\\
   0&  254&      0&      0&        0\\
   0&    0&     22&    279&        0\\
   0&    0&     17&    275&        0\\
   0&    0&      0&      0&      270\\
\end{array} \right)
\]
\caption{Confusion matrix of Hong Kong dataset.}
\label{mat:ConfMatrix_HK}
\end{subfigure}
\hfill
\begin{subfigure}{0.45\textwidth}
\centering
\[
\left( \begin{array}{ccccc}
3514	&	0	&	0	&	0	&	0	\\
0	&	192	&	0	&	0	&	0	\\
0	&	0	&	297	&	0	&	0	\\
0	&	0	&	0	&	285	&	0	\\
0	&	0	&	0	&	0	&	286	\\
\end{array} \right)
\]
\caption{Confusion matrix of London dataset.}
\label{mat:ConfMatrix_LN}
\end{subfigure}
\caption{Confusion matrices for the RTT datasets. The classes are regular, Los Angeles1, Madrid, Moscow, and S\~ao Paulo. Rows correspond to true classes, and columns to predicted classes.}
\label{fig:confusion_matrices_RTT}
\end{figure}

Figure \ref{fig:Mosaic_HK} shows that the regular class is the most frequent in the test set, with the primary misclassification occurring to Los Angeles1. A systematic confusion is observed between the Madrid and Moscow relays, with most observations from both classes classified as Moscow, as indicated by the third and fourth blue-dominated stripes in Figure \ref{fig:Mosaic_HK}. It is likely that this confusion arises due to limited RTT contrast between the two European relays, hindering reliable discrimination. Global outliers are detected across all classes, reflecting the inherent heterogeneity of Internet traffic, with proportions ranging from 1.4\% (regular) to 9.3\% (São Paulo).

\begin{figure}[htbp]
\centering
\begin{subfigure}{0.48\textwidth}
    \centering
    \includegraphics[width=\linewidth]{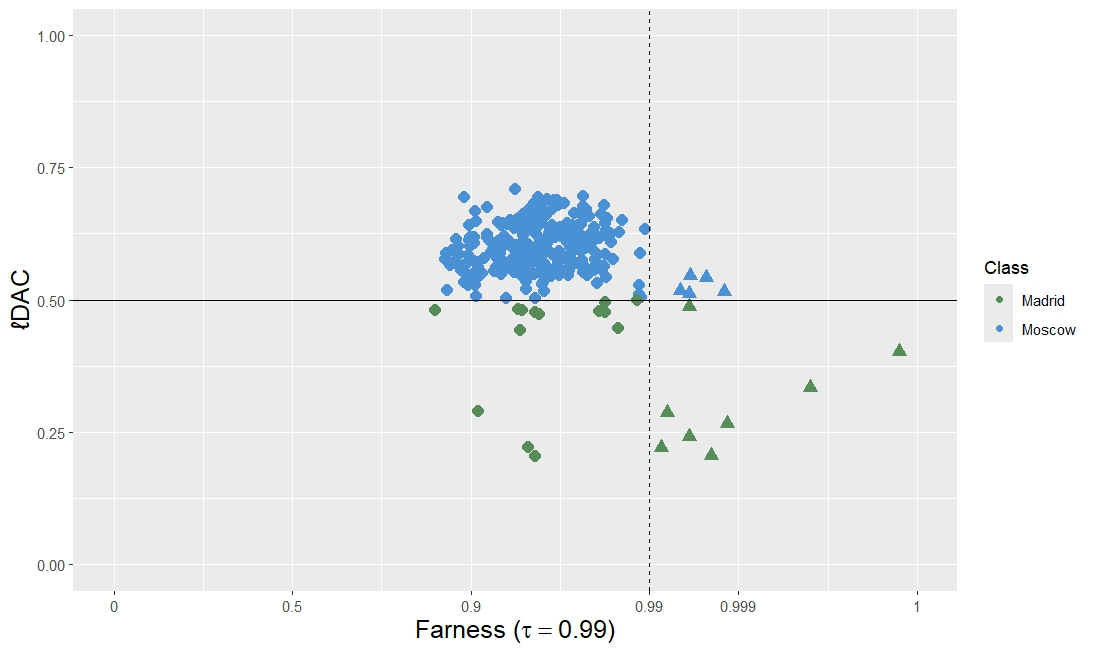}
    \caption{Hong Kong, Madrid relay.}
    \label{fig:PAC_farness_HK_Madrid}
\end{subfigure}
\hfill
\begin{subfigure}{0.48\textwidth}
    \centering
    \includegraphics[width=\linewidth]{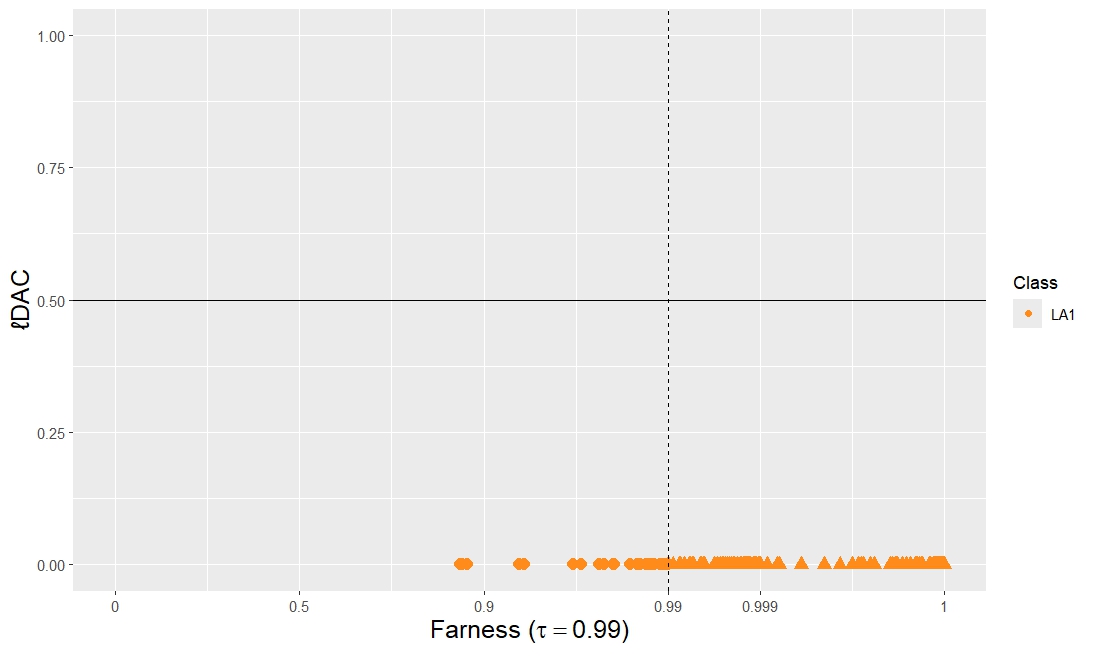}
    \caption{London.}
    \label{fig:PAC_farness_LN_LA1}
\end{subfigure}
\caption{Class maps plots for Madrid class of the Hong Kong dataset and the Los Angeles1 class for the London dataset ($\tau=0.99$). }
\label{fig:Classmap_RTT}
\end{figure}

The class map in Figure \ref{fig:PAC_farness_HK_Madrid} shows that nearly all observations from this class have a ${\rm \ell DAC}$ greater than 0.5. Among the few correctly classified cases, approximately half exhibit uncertainty in their assignments due to their proximity to 0.5. A very high farness value indicates that the test observations deviate substantially from those in the corresponding training class. For traffic redirected from Madrid, 4.3\% of correctly classified observations have farness exceeding 0.99 and are thus flagged as global outliers, highlighting the classifier’s difficulty in discriminating between these two classes.

The London example reveals a pattern not apparent from the confusion matrix. Although the classifier achieves perfect accuracy ($\mathrm{Acc}=1$, see Figure \ref{mat:ConfMatrix_LN}), a large proportion of observations are identified as global outliers: 85.4\% for traffic redirected through Los Angeles1 and 17.1\% for the regular class (see Figure \ref{fig:Mosaic_London} and Figure \ref{fig:PAC_farness_LN_LA1} for details). The corresponding proportions for Madrid, Moscow, and São Paulo1 are 6.1\%, 0.7\%, and 9.8\%, respectively.

\begin{figure}[htbp]
\centering
\begin{subfigure}{0.48\textwidth}
    \centering
    \includegraphics[width=\linewidth]{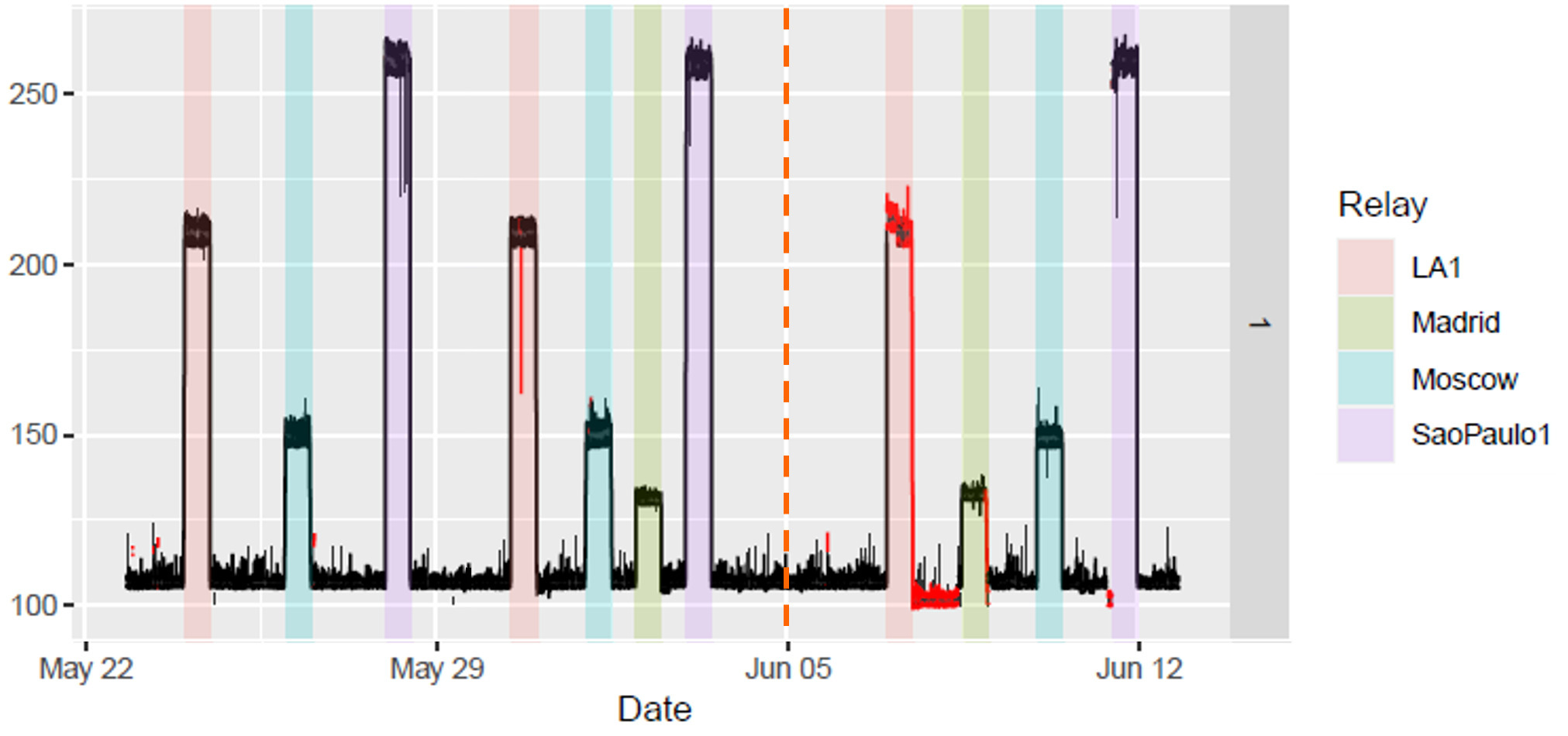}
    \caption{First dimension.}
    \label{fig:Traffic_Dim1_London}
\end{subfigure}
\hfill
\begin{subfigure}{0.48\textwidth}
    \centering
    \includegraphics[width=\linewidth]{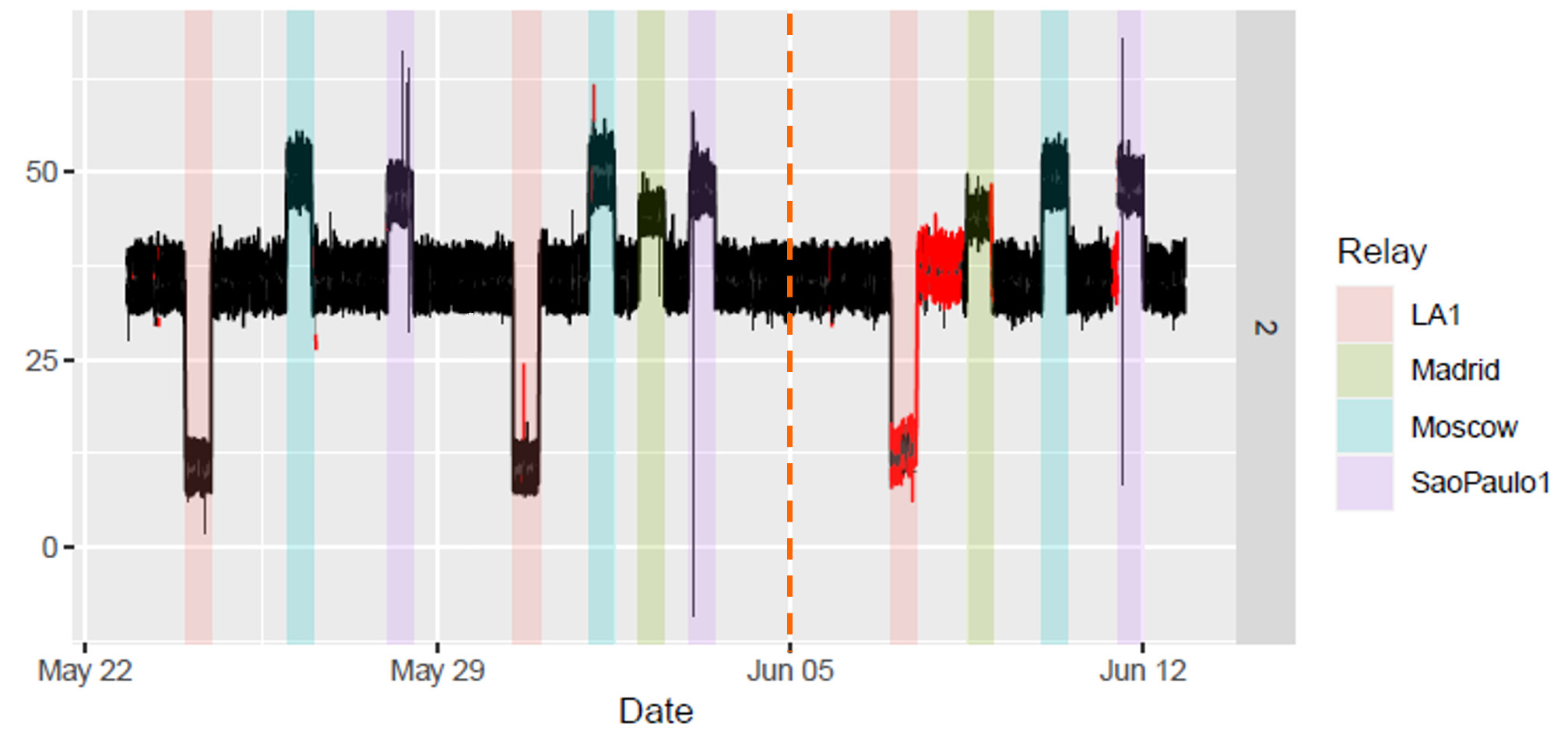}
    \caption{Second dimension.}
    \label{fig:Traffic_Dim2_London}
\end{subfigure}
\caption{London observations projected on the two dimensions suggested by IFDA. In red are observations assigned as global outliers ($\tau=0.99$). The vertical dashed lines separate the training and test sets.}
\label{fig:LondonTraffic}
\end{figure}

By examining the full dataset projected onto the first and second dimensions estimated by IFDA, the underlying behaviour becomes clearer. Figure \ref{fig:LondonTraffic} shows the RTT measurements over time, with the x-axis representing the measurement timestamps. For each timestamp, the RTT interval of observations that were not flagged as global outliers is displayed as a vertical black line, while global outliers are marked in red. The shaded regions denote periods during which redirection attacks occurred, and the vertical dashed lines separate the training and test sets. In Figure \ref{fig:Traffic_Dim1_London} and Figure \ref{fig:Traffic_Dim2_London}, a general decrease in RTT values relative to earlier timestamps is evident, fully explaining this detection by the farness indices. For LA1 (Los Angeles1), although traffic from this class is well separated from the others, the RTT pattern in the test set differs markedly from earlier observations, explaining why nearly all LA1 test data are classified as global outliers.

\section{Conclusions}\label{sec:Concl}

This work introduced the Interval Fisher’s Discriminant Analysis (IFDA), a generalization of Fisher’s classical method to interval-valued data. Building on Moore’s interval arithmetic and Mallows' distance, IFDA extends the Fisher criterion by showing that the total inertia of interval data can be decomposed as the sum of between-class and within-class components.

In the projection space, each component further divides into two complementary sources of variability: one arising from the centres and another from the ranges of the intervals. This decomposition preserves the geometric interpretability of the Fisher approach while explicitly modelling the intrinsic variability and uncertainty inherent to symbolic data.

A numerical optimisation framework was proposed to estimate the discriminant vectors under both orthogonality-constrained and unconstrained settings. The resulting classifier combines interpretability with high predictive performance. Simulation experiments demonstrated that IFDA correctly identifies class structures under varying degrees of overlap and imbalance, while maintaining interpretability through the centre–range decomposition of the Mallows' distance.

The application of the proposed method to the round-trip time (RTT) dataset further confirmed IFDA’s effectiveness. The method successfully detected redirection attacks and uncovered structural traffic patterns that were not discernible from standard confusion matrix analysis.

To enhance interpretability, visualization tools, such as class maps, stacked mosaic plots, and silhouette plots, were adapted to the symbolic framework. These representations provide complementary insights into class separability, uncertainty, and outlier behaviour, extending the analytical power of IFDA beyond numerical summaries. Moreover, the proposed visualization methods are general and can be applied to any classifier based on a distance or dissimilarity measure.

Overall, IFDA offers a coherent and interpretable framework for supervised classification with interval-valued data. By coupling a rigorous theoretical foundation with practical visualization tools, it bridges the gap between symbolic data analysis and classical discriminant methods.

\section*{Acknowledgements}

The authors are grateful to Dr Paulo Salvador and Dr Ana Subtil for providing access to the RTT dataset used in the examples. This work was supported by Fundação para a Ciência e Tecnologia, Portugal, through the projects [\href{https://doi.org/10.54499/UID/PRR/04621/2025}{UID/04621/2025}, UID/4459/2025].

\appendix

\section*{Appendix A. Proof of \eqref{symbbetween} and \eqref{symbwithin}}\label{appendixA}

\begin{align}
   \hspace{-15pt} \mathrm{BI}\ &=\ \nonumber \sum_{j=1}^gn_j\hspace{1pt}d_M(\overline{y}_{j},\overline{y}_B)^2\\ \nonumber
    &=\ \sum_{j=1}^gn_j\Big[(\overline{c}_{y_j} - \overline{c}_{y})^2\ +\ \delta(\overline{r}_{y_j} - \overline{r}_y)^2\Big]\\\nonumber
    &=\ \sum_{j=1}^gn_j\Big[(\boldsymbol{\alpha}^T\overline{\boldsymbol{c}}_{j} - \boldsymbol{\alpha}^T\overline{\boldsymbol{c}})^2\ +\ \delta(|\boldsymbol{\alpha}|^T\overline{\boldsymbol{r}}_{j} - |\boldsymbol{\alpha}|^T\overline{\boldsymbol{r}})^2\Big]\\\nonumber
    &=\ \sum_{j=1}^gn_j\Big[\boldsymbol{\alpha}^T(\overline{\boldsymbol{c}}_{j} - \overline{\boldsymbol{c}})(\overline{\boldsymbol{c}}_{j} - \overline{\boldsymbol{c}})^T\boldsymbol{\alpha}\ +\ \delta|\boldsymbol{\alpha}|^T(\overline{\boldsymbol{r}}_{j} - \overline{\boldsymbol{r}})(\overline{\boldsymbol{r}}_{j} - \overline{\boldsymbol{r}})^T|\boldsymbol{\alpha}|\Big]\\\nonumber
    &=\ \boldsymbol{\alpha}^T\sum_{j=1}^gn_j(\overline{\boldsymbol{c}}_{j} - \overline{\boldsymbol{c}})(\overline{\boldsymbol{c}}_{j} - \overline{\boldsymbol{c}})^T\boldsymbol{\alpha}\ +\ \delta|\boldsymbol{\alpha}|^T\sum_{j=1}^gn_j(\overline{\boldsymbol{r}}_{j} - \overline{\boldsymbol{r}})(\overline{\boldsymbol{r}}_{j} - \overline{\boldsymbol{r}})^T|\boldsymbol{\alpha}|\\
    &=\ \boldsymbol{\alpha}^T\boldsymbol{B}_C\boldsymbol{\alpha}\ +\ \delta |\boldsymbol{\alpha}|^T\boldsymbol{B}_R|\boldsymbol{\alpha}|.
\end{align}
{\allowdisplaybreaks
\begin{align}
    \mathrm{WI} \ &=\ \nonumber\sum_{j=1}^g\sum_{h\in j} d_M(y_h,\overline{y}_{\!j})^2\\\nonumber
    &=\ \sum_{j=1}^g\sum_{h\in j}\Big[(c_{y_h} - \overline{c}_{y_j})^2 + \delta(r_{y_h} - \overline{r}_{y_j})^2\Big]\\\nonumber
    &=\ \sum_{j=1}^g\sum_{h\in j}\Big[(\boldsymbol{\alpha}^T\boldsymbol{c}_{h} - \boldsymbol{\alpha}^T\overline{\boldsymbol{c}}_{j})^2 + \delta(|\boldsymbol{\alpha}|^T\boldsymbol{r}_{h} - |\boldsymbol{\alpha}|^T\overline{\boldsymbol{r}}_{j})^2\Big]\\\nonumber
    &=\ \sum_{j=1}^g\sum_{h\in j}\Big[\boldsymbol{\alpha}^T(\boldsymbol{c}_{h} - \overline{\boldsymbol{c}}_{j})(\boldsymbol{c}_{h} - \overline{\boldsymbol{c}}_{j})^T\boldsymbol{\alpha} + \delta|\boldsymbol{\alpha}|^T(\boldsymbol{r}_{h} - \overline{\boldsymbol{r}}_{j})(\boldsymbol{r}_{h} - \overline{\boldsymbol{r}}_{j})^T|\boldsymbol{\alpha}|\Big]\\\nonumber
    &=\ \boldsymbol{\alpha}^T\sum_{j=1}^g\sum_{h\in j}(\boldsymbol{c}_{h} - \overline{\boldsymbol{c}}_{j})(\boldsymbol{c}_{h} - \overline{\boldsymbol{c}}_{j})^T\boldsymbol{\alpha}\ +\ \delta|\boldsymbol{\alpha}|^T\sum_{j=1}^g\sum_{h\in j}(\boldsymbol{r}_{h} - \overline{\boldsymbol{r}}_{j})(\boldsymbol{r}_{h} - \overline{\boldsymbol{r}}_{j})^T|\boldsymbol{\alpha}|\\
    &=\ \boldsymbol{\alpha}^T\boldsymbol{W}_{\!\!C}\boldsymbol{\alpha}\ +\ \delta|\boldsymbol{\alpha}|^T\boldsymbol{W}_{\!\!R}|\boldsymbol{\alpha}|.
\end{align}

\section*{Appendix B. Proof of {Proposition \ref{lemmaderivatives}}}\label{appendixB}

Let $\boldsymbol{y} = (y_1,\ldots, y_p)^T \in \R^p$. We define the partial derivative of $\boldsymbol{y}$ with respect to the vector $\boldsymbol{\alpha}=(\alpha_1,\ldots,\alpha_p)^T \in \R^p$ as $\dfrac{\partial \boldsymbol{y}}{\partial\boldsymbol{\alpha}} = \begin{bmatrix}
        \dfrac{\partial y_j}{\partial \alpha_i}
    \end{bmatrix}_{i,j=1,\ldots,p}$. 
In order to consider derivatives of the absolute value, we assume that $\alpha_k \neq 0$, $k=1,\ldots,p$.  

To prove (1) in Proposition \eqref{Eq:Prop3.5.1}, it is sufficient to notice that:
\begin{equation}\label{eq:deriv_abs_abs_T_const}
 \frac{\partial (\boldsymbol{b}^T|\boldsymbol{\alpha}|)}{\partial\boldsymbol{\alpha}}\ =\ \begin{bmatrix}
    \dfrac{\partial \left(\sum_{k=1}^pb_k|\alpha_k|\right)}{\partial \alpha_i}
    \end{bmatrix}_{i=1,\ldots,p} \ =\ \mathrm{diag}(\mathrm{sgn}(\boldsymbol{\alpha}))\boldsymbol{b} \hspace{2pt}.
\end{equation}

Let $\boldsymbol{B}$ be a $p\times p$ matrix where $\boldsymbol{B}_{i\boldsymbol{\cdot}}$ denotes its $i$-th row,  $\boldsymbol{B}_{\boldsymbol{\cdot} j}$ its $j$-th column, and $b_{ij}$ the entry in the $i$-th row and $j$-th column.

In the case of (2), we first calculate the partial derivative with respect to $\alpha_i$,
\begin{align}
\frac{\partial (|\boldsymbol{\alpha}|^T\boldsymbol{B}|\boldsymbol{\alpha}|)}{\partial\alpha_i}\ &=\ \frac{\partial}{\partial\alpha_i}\left(\sum_{k=1}^p|\alpha_k|\boldsymbol{B}_{k \boldsymbol{\cdot}}|\boldsymbol{\alpha}|\right)\notag\\ 
&=\ \mathrm{sgn}(\alpha_i)\boldsymbol{B}_{i \boldsymbol{\cdot}}|\boldsymbol{\alpha}|\ +\ \sum_{k=1}^p|\alpha_k|\mathrm{sgn}(\alpha_i)b_{ki}\notag\\ 
&=\ \mathrm{sgn}(\alpha_i)\left(\boldsymbol{B}_{i \boldsymbol{\cdot}} + \boldsymbol{B}_{ \boldsymbol{\cdot}i}^T\right)|\boldsymbol{\alpha}|
\end{align}
from which follows that
\begin{equation}
\label{diffbmatrixabsol}
    \frac{\partial (|\boldsymbol{\alpha}|^T\boldsymbol{B}|\boldsymbol{\alpha}|)}{\partial\boldsymbol{\alpha}}\ =\ \begin{bmatrix}
    \mathrm{sgn}(\alpha_i)\left[\boldsymbol{B}_{i\boldsymbol{\cdot}}|\boldsymbol{\alpha}| + \boldsymbol{B}_{ \boldsymbol{\cdot}i}^T|\boldsymbol{\alpha}|\right]
    \end{bmatrix}_{i=1,\ldots,p} \ =\ \mathrm{diag}\big(\mathrm{sgn}(\boldsymbol{\alpha})\big)(\boldsymbol{B} + \boldsymbol{B}^T)|\boldsymbol{\alpha}|.
\end{equation}

If $\boldsymbol{B}$ is symmetric, then $\boldsymbol{B} = \boldsymbol{B}^T$ and the result (2) follows. 

As to the final identity, we note that the matrices $\boldsymbol{B}_C$, $\boldsymbol{B}_R$, $\boldsymbol{W}_{\!\!C}$, and $\boldsymbol{W}_{\!\!R}$ are symmetric, and apply the derivative of the quotient,
\begin{equation}\label{fprimepre}
\frac{\partial \fun(\boldsymbol{\alpha})}{\partial \boldsymbol{\alpha}}\ =\ \dfrac{\dfrac{\partial \gamma(\boldsymbol{\alpha})}{\partial \boldsymbol{\alpha}}\beta(\boldsymbol{\alpha}) - \gamma(\boldsymbol{\alpha})\dfrac{\partial \beta(\boldsymbol{\alpha})}{\partial \boldsymbol{\alpha}}}{\beta^2(\boldsymbol{\alpha})}\ =\ \frac{1}{\beta(\boldsymbol{\alpha})}\Bigg[\frac{\partial \gamma(\boldsymbol{\alpha})}{\partial \boldsymbol{\alpha}} - \fun(\boldsymbol{\alpha}) \frac{\partial \beta(\boldsymbol{\alpha})}{\partial \boldsymbol{\alpha}}\Bigg],
\end{equation}
\noindent where
\begin{equation}\label{diffgamma}
\frac{\partial \gamma(\boldsymbol{\alpha})}{\partial \boldsymbol{\alpha}}\ =\ \frac{\partial (\boldsymbol{\alpha}^T\boldsymbol{B}_C\boldsymbol{\alpha} + \delta| \boldsymbol{\alpha}|^T\boldsymbol{B}_R|\boldsymbol{\alpha}| )}{\partial \boldsymbol{\alpha}}\ =\ 2\boldsymbol{B}_C\boldsymbol{\alpha} + 2\delta \mathrm{diag}(\mathrm{sgn}(\boldsymbol{\alpha})) \boldsymbol{B}_R |\boldsymbol{\alpha}|, 
\end{equation}
\begin{equation}\label{diffbeta}
\frac{\partial \beta(\boldsymbol{\alpha})}{\partial \boldsymbol{\alpha}}\ =\ \frac{\partial (\boldsymbol{\alpha}^T\boldsymbol{W}_{\!\!C}\boldsymbol{\alpha} + \delta| \boldsymbol{\alpha}|^T\boldsymbol{W}_{\!\!R}|\boldsymbol{\alpha}| )}{\partial \boldsymbol{\alpha}}\ =\ 2\boldsymbol{W}_{\!\!C}\boldsymbol{\alpha} + 2\delta \mathrm{diag}(\mathrm{sgn}(\boldsymbol{\alpha})) \boldsymbol{W}_{\!\!R} |\boldsymbol{\alpha}|.
\end{equation}

By \eqref{diffgamma} and \eqref{diffbeta}, it follows from \eqref{fprimepre} that
\begin{equation}
\label{diffbfisher}
    \frac{\partial \fun(\boldsymbol{\alpha})}{\partial \boldsymbol{\alpha}}\ =\ \frac{2}{\beta(\boldsymbol{\alpha})}\Bigg[\big(\boldsymbol{B}_C - \boldsymbol{W}_{\!\!C} \hspace{1pt}\fun(\boldsymbol{\alpha}) \big)\boldsymbol{\alpha} + \delta \mathrm{diag}(\mathrm{sgn}(\boldsymbol{\alpha})) \big(\boldsymbol{B}_R - \boldsymbol{W}_{\!\!R} \hspace{1pt}\fun(\boldsymbol{\alpha})\big)|\boldsymbol{\alpha}| \Bigg],
\end{equation}
concluding the proof.

\printbibliography

\end{document}